\documentclass[letterpaper]{article} 
\usepackage[draft]{aaai2027} 
\usepackage[hyphens]{url}  
\usepackage{graphicx} 
\def\UrlFont{\rm}  
\usepackage{natbib}  
\usepackage{caption} 
\usepackage{algorithm}
\usepackage{algorithmic}
\usepackage{amsmath}
\usepackage{newfloat}
\usepackage{listings}
\DeclareCaptionStyle{ruled}{labelfont=normalfont,labelsep=colon,strut=off} 
\floatstyle{ruled}
\newfloat{listing}{tb}{lst}{}
\floatname{listing}{Listing}

\usepackage{booktabs}
\usepackage{pdfpages}

\usepackage{multirow}
\usepackage{tabularx}

\title{$S^3$-Bench: Evaluating Speech Interaction Models as Scientific Voice Assistants}

\author{
    Heyang Liu\textsuperscript{1,2}\equalcontrib, Jiayi Huang\textsuperscript{1}\equalcontrib, Wenyang Xiao\textsuperscript{1}, Ziyang Cheng\textsuperscript{1,2}, Lixin Zhang\textsuperscript{1}, Zhen Liu\textsuperscript{1}, Miao He\textsuperscript{3}, Ronghua Wu\textsuperscript{2}, Qunshan Gu\textsuperscript{2}, Yanfeng Wang\textsuperscript{1}, Yu Wang\textsuperscript{1}\corresponding
    \\
}
\affiliations{
    \textsuperscript{\rm 1}Shanghai Jiao Tong University \hspace{10mm}
    \textsuperscript{\rm 2}Ant Group \hspace{10mm}
    \textsuperscript{\rm 3}Tongji University

}

\begin{document}

\maketitle

\begin{abstract}

The advance of multimodal large language models (MLLMs) has fundamentally reshaped the paradigm of human-computer interaction, especially speech interaction models capable of seamless conversations. Despite remarkable performance as general voice assistants, their performance in specialized domains remains underexplored, particularly in scientific areas. Scientific interactions introduce formidable challenges, involving rare technical terminology, spoken norms of abbreviations, and the natural verbalization of symbolic special expressions. In this paper, we introduce S$^3$-Bench, a systematic evaluation framework covering 10 major disciplines, consisting of a Knowledge set for speech question-answering and a Dialogue set for multi-turn progressive interactions with simulated user agents. By decomposing a complete atomic turn into stages of speech recognition, perception, knowledge utilization with reasoning, and response pronunciation, we systematically characterize the common challenges and performance tradeoffs of existing approaches. Furthermore, experiments on multi-turn interactions reveal persistent limitations in user adaptation and the generation of accurate, comprehensive, and efficient responses.

\end{abstract}

\section{Introduction}

The rapid proliferation and advancement of Large Language Models (LLMs) have fundamentally transformed the paradigm of human-computer interaction, which is most prominently exemplified by the emergence of multimodal models equipped with speech interaction capabilities~\cite{wang2025vocalnet, xu2025qwen2, xu2025qwen3, team2025fun}. By integrating specialized encoders and decoders, these architectures achieve seamless, natural, and end-to-end speech generation. Recent evaluations demonstrate that contemporary speech-interactive models exhibit robust performance in generalized scenarios, spanning general knowledge responding, elementary reasoning, creative composition, and multi-turn conversations~\cite{yan2025uro, liu2025vocalbench, li2025televal}. These promising results underscore the immense potential of speech interaction models as intelligent voice assistants and conversational agents.

\begin{figure*}[t]
\centering
\includegraphics[width=0.8\textwidth]{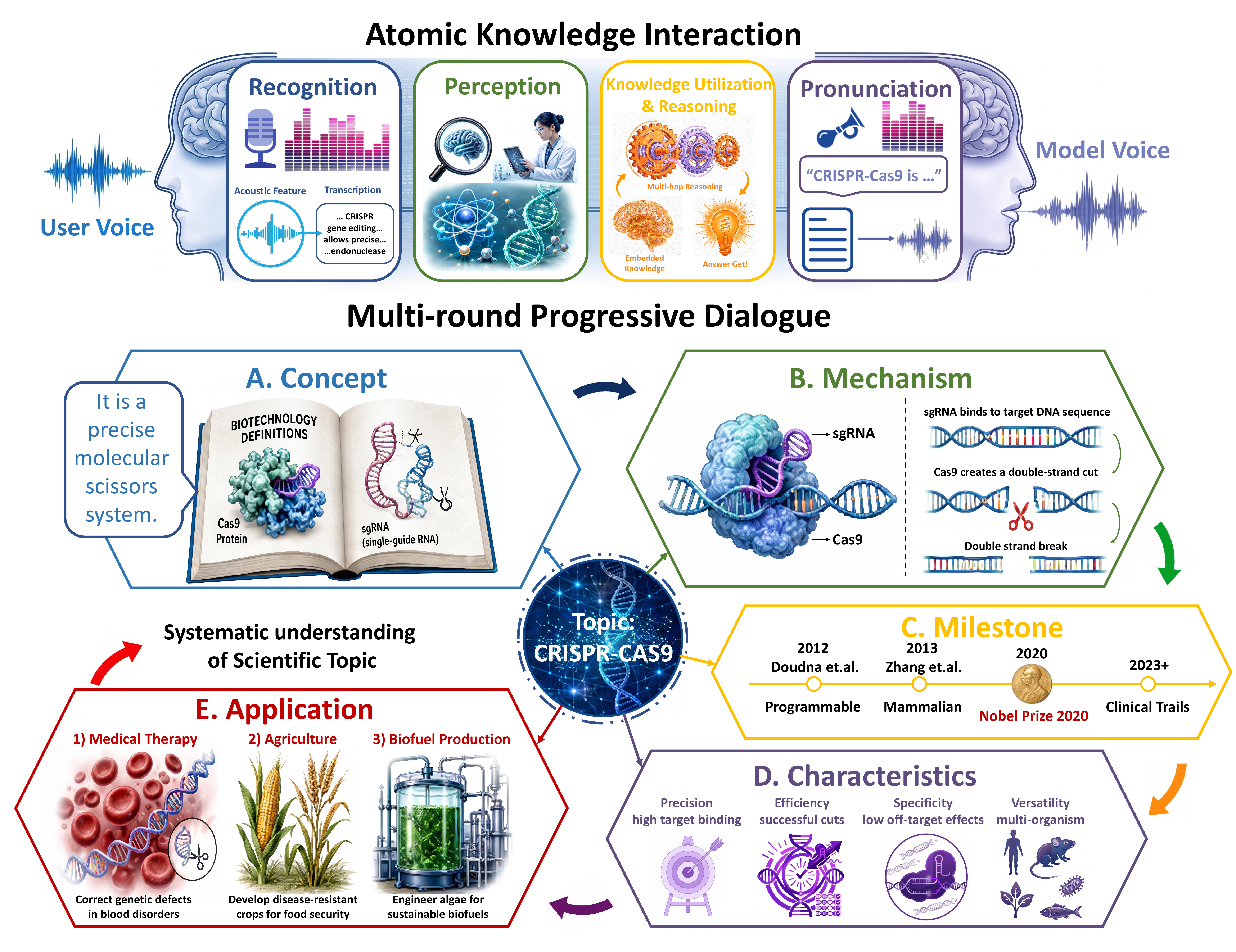} 
\caption{An atomic knowledge question answering in the scientific domain and multi-round progressive dialogue.}
\label{fig:main_fig}
\end{figure*}

Notwithstanding these advancements, a critical question remains: \emph{Can existing speech interaction frameworks sustain their efficacy when deployed in highly complex, scientific conversational settings?} Navigating professional scientific discourse introduces formidable challenges that transcend general speech processing. Specifically, as shown in Figure~\ref{fig:main_fig}, a successful turn of scientific question-answering necessitates that the model accurately recognize key expressions within the user's query, perceive critical acoustic segments to map them to corresponding scientific concepts, leverage its internal knowledge base and reasoning capabilities to derive the correct response, and produce a clear and articulate speech response that follows professional conventions. Beyond generic knowledge-intensive queries, scientific communication plays a critical role in knowledge dissemination and academic discourse. Within this paradigm, models must dynamically discern the user’s foundational knowledge and expertise level, formulate intellectually challenging explanations, and sustain a focus on cutting-edge scientific frontiers. Furthermore, they are required to offer heuristic insights and guidance across multi-turn interactions. By systematically evaluating speech-interactive models in scientific areas, we can elucidate their capabilities and prospects as autonomous science communicators and collaborative research assistants.

In this paper, we introduce the \textbf{S}cientific \textbf{S}peech-to-\textbf{S}peech \textbf{Bench}mark (\textbf{$S^3$-Bench}), a comprehensive evaluation framework spanning 10 major scientific domains alongside their respective sub-fields. $S^3$-Bench comprises a specialized college-level scientific question-answering subset ($S^3$-Knowledge) and a multi-turn dialogue collection ($S^3$-Dialogue). The former specifically targets professional terminology, colloquial abbreviations, and domain-specific linguistic expressions. By decoupling the spoken response process into Recognition, Perception, Knowledge Utilization \& Reasoning, Generation \& Pronunciation, we systematically expose the pervasive limitations of mainstream speech models in scientific contexts. For the dialogue benchmark, we construct multi-turn, progressively advancing dynamic conversations structured around frontier sciences and cutting-edge techniques. By simulating diverse user profiles with varying personas and academic backgrounds, we comprehensively evaluate model performance across Scientific Factuality, Audience Adaptation, and Inquiry Efficiency. Our contributions can be summarized as follows:

\begin{itemize}
    \item We propose $S^3$-Bench, the first comprehensive evaluation framework dedicated specifically to scientific interaction scenarios. 
    \item We systematically decouple the atomic interaction into four distinct, logical stages, and successfully uncover the pervasive causes of performance bottlenecks in contemporary speech interaction models.
    \item By constructing progressively advancing, multi-turn scientific dialogues embedded with diverse user personas, we validate the actual adaptation limits and deployment potential as scientific voice assistants.
\end{itemize}




\section{$S^3$-Bench}
\label{sec:data_pipeline}


\begin{figure*}[t]
    \centering
    \includegraphics[width=0.88\linewidth]{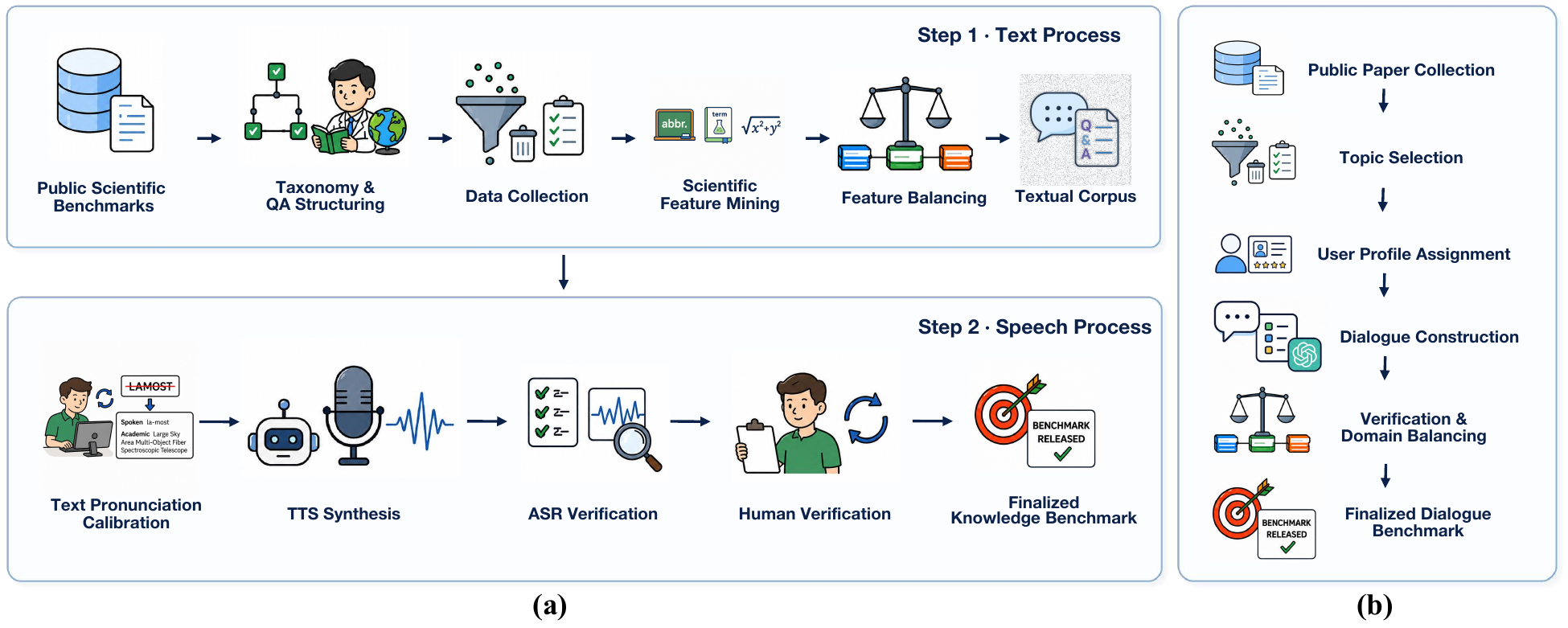}
    \caption{Benchmark construction pipeline for (a) $S^3$-Knowledge and (b) $S^3$-Dialogue.}
    \label{fig:pipeline}
\end{figure*}

\begin{figure*}[h]
    \centering
    \includegraphics[width=0.95\linewidth]{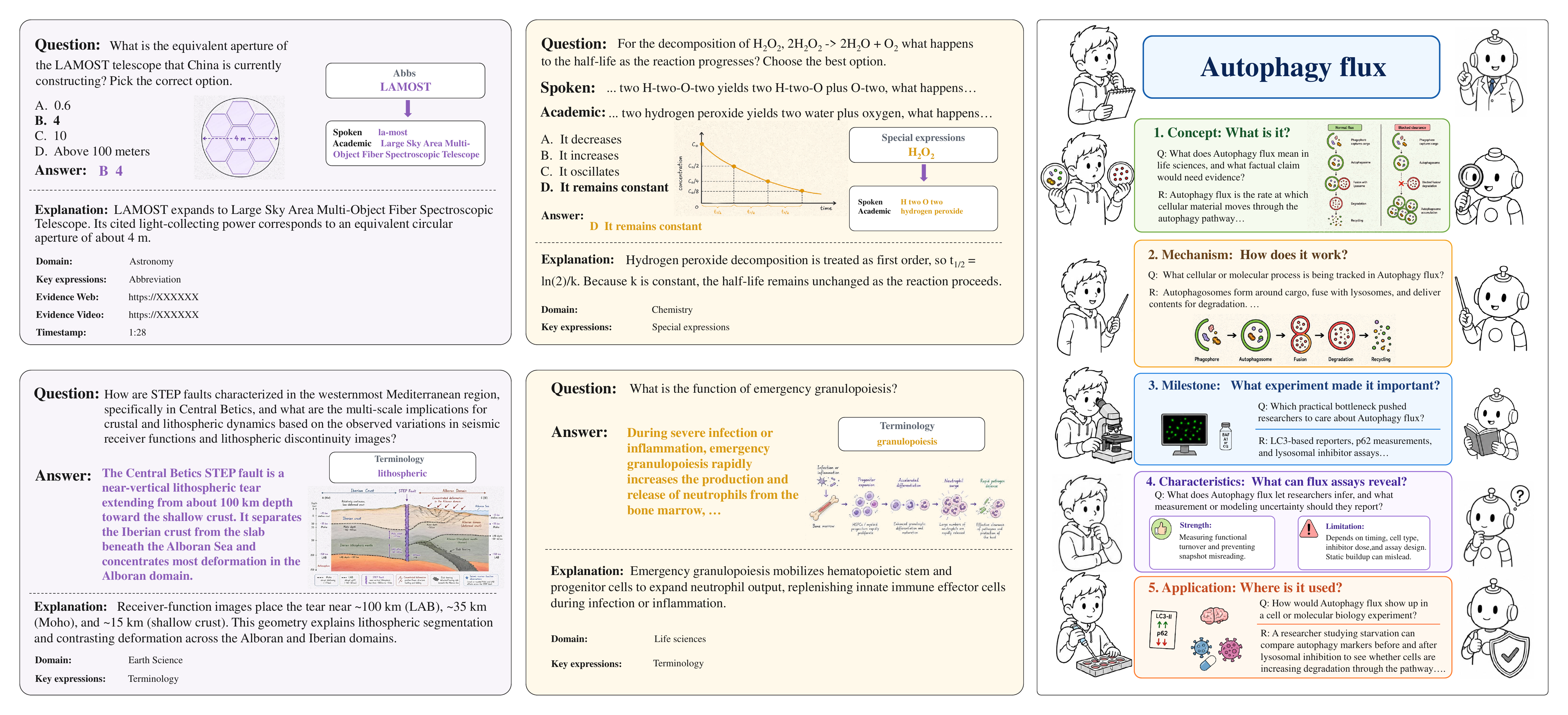}
    \caption{Benchmark examples in $S^3$-Bench.}
    \label{fig:benchmark_exmaples}
\end{figure*}

\begin{figure*}[h!]
    \centering
    \includegraphics[width=0.85\linewidth]{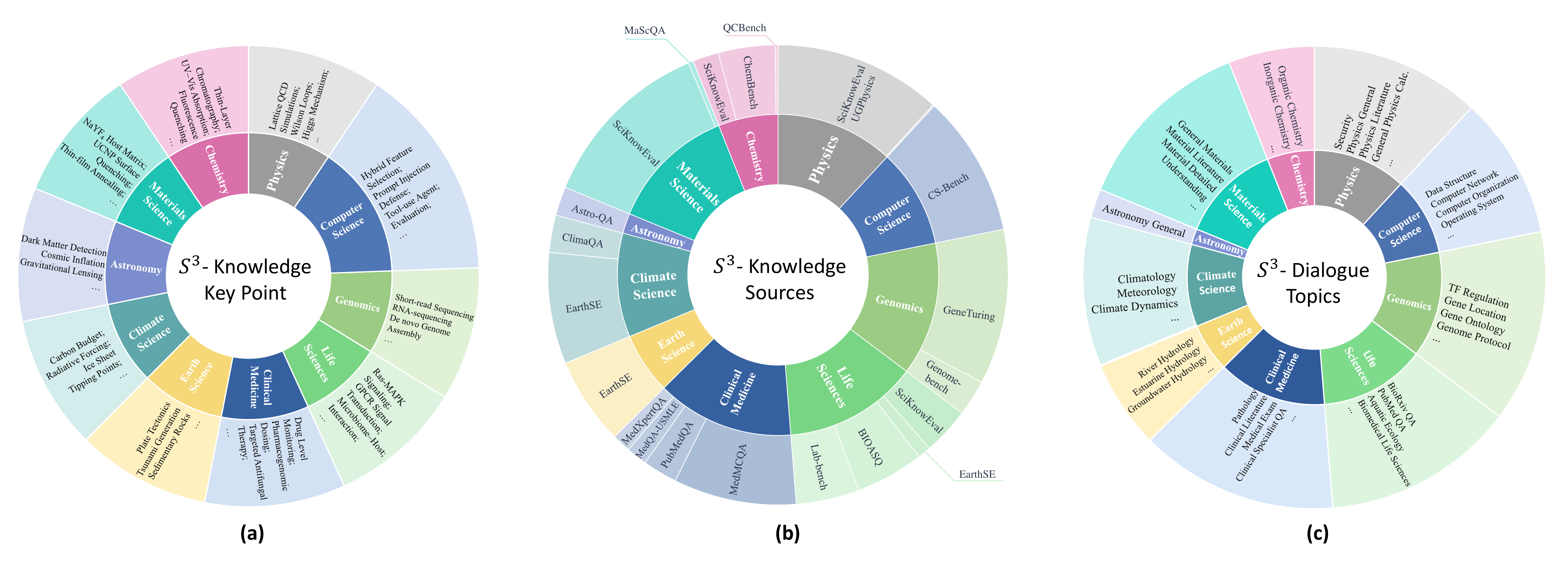}
    \caption{Dataset Statistics. (a) $S^3$-Knowledge key points in various domains; (b) $S^3$-Knowledge data source distribution; (c) $S^3$-Dialogue domain distribution and included topics.}
    \label{fig:benchmark_statisctics}
\end{figure*}

\subsection{Scientific Speech Interaction}

\subsubsection{Key Scientific Expressions}

Spoken interactions within scientific domains impose multifaceted challenges on multimodal models, predominantly manifesting through complex terminology, diverse abbreviations, and specialized expressions. Scientific terminology suffers from the inherent data sparsity of these professional terms in mainstream pre-training corpora, and frequently acts as low-frequency bottlenecks that hinder acoustic perception and semantic grounding. In addition, abbreviations often possess conventionalized, domain-specific pronunciations rather than a naive mapping of individual letter spellings, widely exhibiting implicit syllable insertions or deletions based on community norms (e.g., acronym BRCA articulated as /\textnormal{'}br\ae k\textnormal{\reflectbox{e}}/). Finally, scientific discourse is heavily interspersed with discipline-specific expressions, including chemical formulas, mathematical equations, truncated genomic sequences, and complex measurement units. Transforming these highly structured textual notations into natural, colloquial acoustic streams demands an exceptionally advanced capacity for contextual reasoning and fluid multimodal alignment.

\subsubsection{Atomic Interactive Turn Process} \label{subsec:knowledge_qa_turn}


A complete speech-interactive turn within the scientific domain encompasses four sequential stages. \textbf{Recognition:} The model aligns the user query speech with their corresponding textual targets, achieving high-fidelity speech recognition. \textbf{Perception:} Moving beyond raw transcription, the model must associate critical substantive words with their corresponding scientific concepts. \textbf{Knowledge Utilization \& Reasoning:} Leveraging its internalized parametric knowledge and underlying cognitive reasoning capabilities, the model synthesizes the acquired information to derive a scientifically accurate answer and formulate a coherent textual response. \textbf{Generation \& Pronunciation:} The model generates an intelligible and audience-facing spoken response. It must ensure that specialized terminology, colloquial abbreviations, and domain-specific linguistic expressions are articulated with precise pronunciation, as they carry critical semantic weight. 



\subsection{Construction Pipeline}

\subsubsection{S$^3$-Knowledge}
As shown in Figure~\ref{fig:pipeline} (a), the textual source data for $S^3$-Knowledge is primarily derived from existing text-based scientific knowledge benchmarks. To guarantee appropriate intellectual difficulty and maintain a rigorous multi-tier disciplinary taxonomy, we predominantly select college-level and domain-specific evaluation datasets. To ensure the dataset's compatibility with spoken interaction, we implement a rigorous data pruning process. Specifically, entries that are excessively lengthy or inherently unsuitable for acoustic modalities are systematically removed. To preserve distinct scientific acoustic characteristics, we adapt the definition of \emph{Terminology} from LaSR~\cite{liu2026lasr}, \emph{Abbreviations} from a regularized string matching algorithm, and \emph{Special Expressions} of specific academic disciplines from manual filtering. Those devoid of these critical scientific features are discarded, and we balance the distribution across different subjects, resulting in a curated corpus of approximately 2,200 high-quality textual entries.

Prior to speech synthesis, substantial manual effort is dedicated to calibrating the pronunciation of specialized vocabularies, with a particular focus on distinguishing between initialisms and acronyms for abbreviations and pronunciation norms of domain-specific symbolic expressions. Given that colloquial academic discourse often introduces pronunciation variances, we cross-reference ambiguous technical terms with online videos and authoritative disciplinary texts to assign standard textual representations optimized for text-to-speech (TTS) models. The corresponding source URLs and timestamps are meticulously preserved to serve as pronunciation verifications. For the audio generation pipeline, we utilize Qwen3-TTS~\cite{hu2026qwen3} to perform zero-shot speech synthesis. To guarantee high audio quality and acoustic diversity, the reference prompts are randomly sampled from the Common Voice test set, strictly filtered to ensure a DNSMOS Pro score~\cite{cumlin2024dnsmos} exceeding 3.5. Following synthesis, Qwen3-ASR and its variants fine-tuned on scientific domains are deployed to transcribe the generated audio~\cite{shi2026qwen3}. Any discrepancies between the transcriptions and the ground-truth texts trigger a manual re-inspection and re-synthesis loop. During this rigorous filtering and refinement process, entries with unfounded pronunciations or synthetic errors are pruned, culminating in a finalized evaluation set of 1,980 high-quality speech samples. Manual verification against professional online dictionaries and authoritative video references indicates that 99\% of the entries exhibit pristine clarity and phonetic accuracy.





\subsubsection{S$^3$-Dialogue}


As shown in Figure~\ref{fig:pipeline} (b), we obtain diverse leading areas and topics from a large-scale collection of research articles~\cite{qin2026data}, and filter representative concepts before transforming them into structured dialogues. Each topic is paired with a synthetic user profile defined by an expertise level and a character trait. The former is divided into Beginner, Undergraduate, Graduate, and Expert, representing increasing requirements for technical depth and methodological details. The character traits are curious, patient, anxious, overconfident, and practical, which affect the framing and communicative intent of the questions. The multi-round dialogue contains progressive atomic queries related to \textbf{Concept}, \textbf{Mechanism}, \textbf{Milestone}, \textbf{Characteristics}, and \textbf{Application}, which are generated by GPT-5.5 Pro and cross-validated by Gemini3.5-flash and Qwen3.7-Max~\cite{qwen37}.


\subsection{Benchmark Statistics} 

The instance examples are illustrated in Figure~\ref{fig:benchmark_exmaples}. The benchmark covers critical scientific areas across 10 domains, as shown in Figure~\ref{fig:benchmark_statisctics}. S$^3$-Knowledge contains 1,980 simulated user queries, partitioned into multiple-choice question answering (MCQA) and open-ended inquiries. The preserved instances are sourced from 17 public evaluation datasets, which ensures the balance and diversity~\cite{feng2025sciknoweval, xu2025earthse, shang2025benchmarking, song2025cs, pal2022medmcqa}. For $S^3$-Dialogue, we endeavor to maintain a balanced distribution across academic disciplines, ultimately retaining 213 multi-turn dialogues. Detailed statistics are shown in Appendix 4.

\section{$S^3$-Knowledge Experiments}

\subsection{Experiment Setup}

\subsubsection{Evaluated Models}


Specifically, six Speech LLMs and two Omni-LLMs are benchmarked, most of which incorporate a standard 7–9B parameter LLM backbone~\cite{wang2025vocalnet,long2026vita, ding2025kimi, zhang2025mimo, wu2025step, team2025fun, xu2025qwen2}, except Qwen3-Omni with a scaled 30B parameter MoE-based Thinker-Talker architecture~\cite{xu2025qwen3}. We also include the Qwen3.5-Omni-flash API~\cite{team2026qwen35}, and a cascaded pipeline linking Whisper-large-v3, Qwen3-8B, and Qwen3-TTS~\cite{radford2023robust, yang2025qwen3, hu2026qwen3}. Detailed model configurations are provided in Appendix 5.

\subsubsection{Evaluation Indicators}
The evaluation metrics in this study are primarily derived from ASR tasks and standard accuracy. Following the conventions of Spoken Darwin-Science, lexical items with a frequency of less than 10 occurrences in GigaSpeech~\cite{chen2021gigaspeech} are designated as terminologies, and their performance is quantified using the Entity Error Rate (EER)~\cite{liu2026lasr}. Furthermore, abbreviations are characterized by consecutive alphanumeric sequences with non-initial uppercase letters and extracted via rule-based methods. Concurrently, the Abbreviation Error Rate (AER) is defined to evaluate the discrepancy. The loose AER eliminates formatting discrepancies by standardization to uppercase and removing spaces (e.g., transforming "d n a" to "DNA"), while the strict AER demands precise case-sensitive and format-compliant abbreviation recognition (e.g., accepting "mRNA" but rejecting "MRNA"). For general performance, we report the word error rate (WER).

To evaluate perceptual capabilities, we introduce the Relevant Rate (RR) to quantify how effectively a model maps critical speech cues to their respective semantic concepts. This metric computes the ratio of responses containing highly relevant key content words during open-ended generation. For standard accuracy, regarding MCQA formats, option indices are first extracted using a deterministic string-matching algorithm. Where direct extraction fails, as well as for open-ended questions, Qwen3.7-Plus is seamlessly integrated to adjudicate the correctness of the responses.

\subsubsection{Fine-tuned ASR Models}

\begin{table}[t!]
\centering
\resizebox{0.5\textwidth}{!}{
\begin{tabular}{ccccccccccccccccccccc}
\toprule

 \multirow{2}{*}{\textbf{Model}} & \multicolumn{3}{c}{\textbf{Terminology EER (\%)}} & \multicolumn{2}{c}{\textbf{Abbs AER (\%)}} \\  \cmidrule(lr){2-4} \cmidrule(lr){5-6}

 & Base & Hard & Avg & Loose & Strict \\ \midrule
Whisper-large-v3 & \textbf{8.35} & 30.56 &  16.86 & \textbf{2.59} & 24.08 \\
Qwen3-ASR-0.6B & 18.50 & 42.93 & 27.84 & 5.52 & 33.20 \\
Qwen3-ASR-1.7B & 10.55 & 30.93 & 18.37 & 3.83 & 28.36 \\ \hline
Qwen3-ASR-Terminology & 8.98 & \textbf{25.63} & \textbf{15.35} & - & - \\
Qwen3-ASR-Abbs & - & - & - & 2.72 & \textbf{16.45}  \\

\bottomrule
\end{tabular}
}
\caption{The performance of ASR models used for evaluation.}
\label{tab:qwen3_asr_performance}
\end{table}


For speech generative capabilities, additional ASR models are necessary to convert the speech responses into text. However, most open-source models underperform on terminologies and abbreviations. The performance on specific test sets is shown in Table~\ref{tab:qwen3_asr_performance}. The statistics for the terminology EER are sourced from the LaSR test set~\cite{liu2026lasr}, while the Abbs AER is benchmarked on real data obtained from YouTube. For further accuracy optimization, we leverage Qwen3-ASR-1.7B~\cite{shi2026qwen3} fine-tuned on Spoken Darwin-Science. We also extract abbreviation-rich sentences from the same textual resources~\cite{qin2026data}. To construct the speech synthesis training set, a comprehensive pronunciation lexicon spanning around 230K abbreviations is established by combining automated English phonetic regularization rules~\cite{enwl_wordlist_2026} with manual refinement. We denote the model fine-tuned on this corpus as Qwen3-ASR-Abbs. The significant decline in strict AER substantiates that the model successfully capitalizes on contextual semantics to differentiate between standard words and abbreviations. The process for constructing the abbreviation training corpus and its distribution is detailed in Appendix 6. For experimental results, the Qwen3-ASR-1.7B model is employed for standard WER. Meanwhile, evaluation on the terminology and abbreviation is performed using the above-mentioned fine-tuned variants.

\subsection{Performance Analysis (Table~\ref{tab:overall_performance})}

\begin{table*}[htp]
\centering
\resizebox{1\textwidth}{!}{
\begin{tabular}{cccccccccccccccccccccc}
\toprule

 \multirow{3}{*}{\textbf{Model}} & \multicolumn{3}{c}{\textbf{Recognition (R)}} & \multicolumn{1}{c}{\textbf{Perception (P)}} & \multicolumn{4}{c}{\textbf{Knowledge Utilization \& Reasoning (K)}} & \multicolumn{3}{c}{\textbf{Generation \& Pronunciation (G)}}  \\  \cmidrule(lr){2-4} \cmidrule(lr){5-5} \cmidrule(lr){6-9}  \cmidrule(lr){10-12}

 & WER (\%) & EER (\%) & AER (\%) & RR (\%) & Ter (\%) & Abb (\%) & SE (\%) & Avg (\%) & WER (\%) & EER (\%) & AER (\%) \\ \midrule

VocalNet & - & - & - & 87.37 & 56.16 & \textbf{56.53} & 55.31 & 56.21 & 4.38 & 22.64 & 20.64 / 34.00   \\
VITA-Audio & 16.38 & 66.26 & 46.60 / 99.96 & 78.48 & 40.80 & 37.87 & 41.34 & 38.86 & 5.31 & 34.70 & 38.72 / 51.05  \\
Kimi-Audio & \textbf{9.62} & \textbf{36.22} & \textbf{18.15} / 34.53  & 94.34 & 56.81 & 49.04 & 50.66 & 50.44 &  20.34 & 47.53 & 39.37 / 47.72 \\
MiMo-Audio & 19.01 & 50.65 & 37.34 / 88.70   & 95.51 & 40.97 & 42.69 & 46.09 & 43.03 & 45.81 & 47.39 & 20.55 / 30.70  \\
Step-Audio-2-mini & 12.07 & 48.73 & 25.43 / 69.06  & 85.00 & 52.90 & 48.50 & 43.64 & 47.96 &  7.63 & 22.41 & 21.36 / 31.42 \\ 
Fun-Audio-Chat & 13.66 & 42.69 & 25.40 / 59.20  & 96.26 & \textbf{62.18} & 56.22 & 58.10 & \textbf{57.58} &  6.68 & \textbf{19.31} & 9.45 / 24.81   \\
Qwen2.5-Omni & 14.94 & 40.01 & 24.46 / 44.40  & 88.69 & 41.83 & 41.99 & 46.37 & 42.53 & \textbf{4.31} & 19.96 & \textbf{9.36} / \textbf{24.71}  \\
Qwen3-Omni &  15.78 & 41.09 & 30.05 / 46.61  & \textbf{97.42} & 53.87 & 50.78 & \textbf{60.89} & 53.18 & 50.24 & 56.63 & 24.67 / 43.16 \\
\hline
Cascade (Qwen3-8B) & 9.48 & 44.36 & 21.05 / 34.44 & 94.85 & 63.04 & 58.16 & 65.92 & 60.35 &  30.91 & 45.69 & 16.54 / 36.38  \\
Qwen3.5-Omni-Flash & \textbf{8.42} & \textbf{33.86} &18.59 / \textbf{31.20}  & 93.79 & \textbf{68.48}  & \textbf{61.59}  & \textbf{66.20}  & \textbf{63.62} &\textbf{3.65} &\textbf{13.30}&17.59 / 35.18 \\
\bottomrule
\end{tabular}
}
\caption{Model performance on $S^3$-Knowledge. \textbf{Bold} indicates the optimal result within offline end-to-end models. For the cascade system and API, results are highlighted in \textbf{Bold} if surpasses all other models. The missing data for VocalNet stems from a lack of task-specific training, causing ASR instruction failures. }
\label{tab:overall_performance}
\end{table*}



\paragraph{Recognition}
Recognition captures whether a model correctly transcribes the keywords in the user query, a prerequisite for downstream comprehension, reasoning, and response. Most models attain comparable query-WER, with Qwen3.5-Omni-Flash (8.42\%) and Kimi-Audio (9.62\%) the strongest. The spread is moderate, indicating that mainstream models already transcribe general queries reasonably well. Errors are concentrated on proper scientific terminologies and abbreviations, especially the former; the best offline model still maintains an error rate of over 36\%, indicating a general challenge. Furthermore, approximately half of the evaluated models exhibit over double AER in strict mode than in loose mode, indicating that accurately recognizing abbreviated expressions and mapping them to their corresponding canonical text patterns remains beyond their capability.

\paragraph{Perception}
Perception measures whether the model associates the salient fragments in the user's speech with the intended scientific concept. The relevant rate is generally high (RR $>$ 85\% for most models). Qwen3-Omni and Fun-Audio-Chat-8B reach 97.42\% and 96.26\% respectively, indicating that once a model hears the relevant words, it can usually paraphrase the associated terms. However, the perception metric consistently exceeds knowledge utilization performance, demonstrating that being able to restate a concept does not equate to grasping its reasoning. Perception thus acts as a threshold rather than a guarantee.

\paragraph{Knowledge Utilization and Reasoning}
This stage, quantified by the accuracy on text responses faced with complex scientific queries, is where models are truly differentiated. Qwen3.5-Omni-Flash leads with an average accuracy of 63.62\%, followed by Cascade (60.35\%), Fun-Audio-Chat-8B (57.58\%), and VocalNet (56.21\%). The failure to recognize terminology and abbreviations severely disrupts the models. While the closed-source interactive API maintains relatively robust performance, among the offline models, only Fun-Audio-Chat achieves an accuracy rate exceeding 60\% on terminology. Special expressions (mathematical symbols, chemical formulae) are also a difficult category. Even the leading API attains 66.20\%, and around half of the offline models fall below 50\%, indicating that while the text backbone can produce structured content, this capability does not transfer smoothly to the speech channel. All models demonstrate varying performance degradation relative to their LLM backbones, with details provided in Appendix 7.

\paragraph{Generation and Pronunciation.}
Generation and pronunciation performance is the most discriminative and most easily obscured stage for users. VocalNet, Qwen2.5-Omni, and Qwen3.5-Omni-Flash achieve WER below 6\% with low EER/AER. On the other hand, models represented by Kimi-Audio, MiMo-Audio, and Qwen3-Omni struggle to handle the pronunciation of complex terminologies, exhibiting notable deficiencies in articulation clarity and highlighting the necessity of overall performance improvement. An instance-level analysis reveals that 84.2\% of correctly answered turns carry special-symbol tokens. The advanced LLM backbone tends to emit highly structured outputs while the speech generation phase cannot render, collapsing overall intelligibility. 

\subsection{Capability Transfer and Consistency Analysis}

\begin{table}[htp]
\centering
\resizebox{0.45\textwidth}{!}{
\begin{tabular}{cccccccc}
\toprule
\textbf{Correlation} & \textbf{Spearman} & \textbf{Pearson} & \textbf{$r^{2}$} & \textbf{Shrinkage $r$}  \\
\midrule
\multicolumn{5}{l}{\textit{Front-three-stage capability chain}} \\
R$\rightarrow$P & 0.393 & 0.325 & 0.11 & 0.361   \\
P$\rightarrow$K & \textbf{0.786} & \textbf{0.747} & \textbf{0.56} & \textbf{0.788}  \\
R$\rightarrow$K & 0.679 & 0.448 & 0.20 & 0.493  \\
\midrule
\multicolumn{5}{l}{\textit{Coupling with the Generation stage}} \\
K$\rightarrow$G & $-0.190$ & $-0.026$ & 0.00 & $-0.028$  \\
P$\rightarrow$G & $-0.595$ & $-0.671$ & 0.45 & $-0.708$   \\
R$\rightarrow$G & $-0.429$ & $-0.238$ & 0.06 & $-0.266$   \\
\bottomrule
\end{tabular}
}
\caption{Cross-stage model-level correlation analysis.}
\label{tab:consistency}
\end{table}

As shown in Table~\ref{tab:consistency}, we conduct a cross-stage model-level correlation analysis on $S^3$-Knowledge. All stages are normalized to higher-is-better accuracy: $R{=}1{-}\mathrm{WER}$, $P{\in}[0,1]$, $K{\in}[0,1]$, $G{=}1{-}\min(1,\mathrm{spoken\text{-}WER})$. Each pair uses the maximal offline end-to-end model subset for which both stages are available. Spearman $\rho$ and Pearson $r$ are reported with $r^{2}$; Shrinkage~$r$ is the Olkin--Pratt small-sample correction of Pearson (unbiased for $n{\le}8$). The front three stages form a positively coupled capability chain ($\rho_{PK}{=}0.786$, $\rho_{RK}{=}0.679$), whereas Generation couples negatively with recognition and perception stage ($\rho_{PG}{=}{-}0.595,\ \rho_{RG}{=}{-}0.429$), evidencing the trade-off. The front three stages are positively coupled across models, reinforcing one another along a coherent capability chain, and this coupling is strongest at perception, promoting knowledge utilization and reasoning. That is, a model that perceives the salient scientific concept is overwhelmingly likely to reason it through correctly. Recognition, in turn, promotes knowledge, confirming that recognition quality is a genuine prerequisite for knowledge accuracy. Yet the link from recognition to perception is markedly weaker ($\rho_{RP}{=}0.393$, $r^{2}{=}0.11$), revealing that perception does not completely rely on word-level transcription. Even when a keyword is mis-recognized, the model recovers the underlying concept from the surrounding context, exhibiting strong contextual reasoning capabilities to mitigate the impact of erroneous transcription on subsequent conversations.


For speech generation, perception and recognition exhibit noticeable negative coupling with the spoken response, reflecting a representation-pronunciation trade-off: models with stronger perceptual capabilities capture denser domain-specific terminology and abbreviations, which downstream vocoders struggle to articulate intelligibly. Crucially, knowledge reasoning and generation demonstrate near-complete statistical independence, confirming that high-level parametric reasoning neither directly hinders nor promotes acoustic generation. Overall, speech generation serves as the primary bottleneck where the capability chain breaks, highlighting the need for dedicated acoustic-textual alignment rather than relying solely on upstream text-reasoning scaling.

\section{$S^3$-Dialogue Experiments}

\begin{table*}[htp]
\centering
\resizebox{1\textwidth}{!}{
\begin{tabular}{cccccccccccccccccccccc}
\toprule

 \multirow{3}{*}{\textbf{Model}} & \multicolumn{6}{c}{\textbf{Scientific Factuality (\%)}} & \multicolumn{1}{c}{\textbf{Audience Adaptation}} & \multicolumn{4}{c}{\textbf{Inquiry Efficiency}}  \\  \cmidrule(lr){2-7} \cmidrule(lr){8-8} \cmidrule(lr){9-12} 

 & B\&L & B\&S & H\&L & H\&S & Avg L & Avg S & AAR (\%) & $\leq$ 5-Turn (\%) & $\leq$ 8-Turn (\%) & $\leq$ 10-Turn (\%)  & Avg Turn  \\ \midrule

VocalNet &77.78 &21.67 &69.33 &15.43 &73.62 &18.59 &45.54 & 4.69 & 22.07 & 40.85 & 10.75   \\
VITA-Audio  &72.78 &14.44 &71.43 &15.81 &72.11 &15.12 &51.17 & 5.16 & 30.05 & 48.83 & 10.29  \\
Kimi-Audio &67.96 &14.63 &65.58 &15.77 &66.79 &15.19 &45.75 & 0.94 & 9.86 & 26.76 & 11.51  \\
MiMo-Audio &69.44 &20.74 &56.00 &13.14 &62.82 &17.00 &50.23 & 14.55 & 45.54 & 70.89 & 8.99 \\
Step-Audio-2-mini &68.15 &10.56 &62.72 &8.16 &65.50 &9.38 &52.13 & 6.64 & 31.28 & 50.24 & 10.21  \\
Fun-Audio-Chat  &87.41 &34.07 &86.29 &32.76 &86.85 &33.43 &53.05 & 28.17 & 80.75 & 95.77 & 6.86 \\
Qwen2.5-Omni &82.96 &20.00 &84.38 &22.10 &83.66 &21.03 &50.23 & 13.62 & 36.62 & 59.62 & 9.52  \\
Qwen3-Omni & \textbf{94.81} & \textbf{47.96} & \textbf{94.86} & \textbf{54.10} & \textbf{94.84} & \textbf{50.99} & \textbf{53.52} & \textbf{55.87} & \textbf{95.77} & \textbf{99.53} & \textbf{5.85}  \\
\hline
Cascade  &92.96 &39.07 &92.19 &34.48 &92.58 &36.81 &\textbf{63.38} & 40.38 & 77.00 & 92.02 & 6.87 \\
Qwen3.5-Omni-Flash &94.07 &40.74 &93.90 &43.81 &93.99 & 42.25 & 55.87 &  \textbf{61.03} & \textbf{95.77} & \textbf{99.53} & \textbf{5.74}  \\
\bottomrule
\end{tabular}
}
\caption{Model performance on $S^3$-Dialogue. Base (B) and Hard (H) refer to difficulty. Loose (L) and Strict (S) refer to criteria.}
\label{tab:dialogue_bench}
\end{table*}

\subsection{Evaluation Methods}

\begin{figure}[t!]
  \centering
  \includegraphics[width=0.4\textwidth]{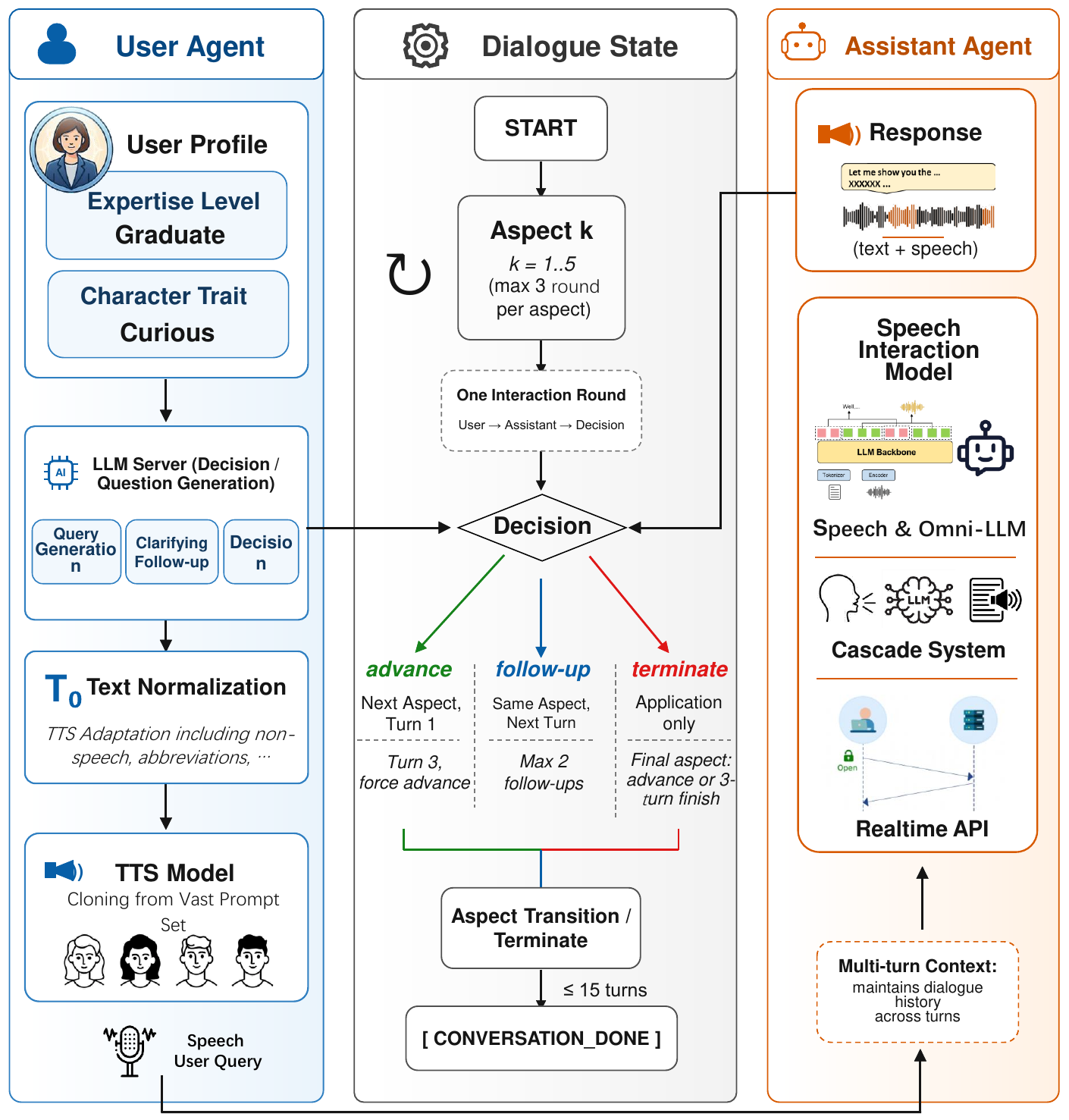}
  \caption{The evaluation pipeline for $S^3$-Dialogue.}
  \label{fig:Dialogue}
\end{figure}

As shown in Figure~\ref{fig:Dialogue}, dialogue performance is evaluated in a multi-turn interaction scenario involving dual agents. The User agent consists of Qwen3.7-Plus~\cite{qwen37} and Qwen3-TTS~\cite{hu2026qwen3}, with an intermediate text normalization module integrated for speech synthesis. High-quality speech prompts are randomly selected from the Common Voice test set~\cite{ardila2020common}. The generated speech queries, along with their corresponding conversational histories, are provided to the speech interaction model to produce aligned textual and spoken responses. The User agent evaluates the assistant reply: for those that successfully address the substantive issue, the agent proceeds to the next aspect until the conversation terminates; for ambiguous or problematic responses, it opts to follow up for logical clarification, with a maximum of three dialogue turns allowed per aspect.

For comprehensive evaluation, we employ a multi-dimensional framework comprising three metrics. Specifically, \textbf{Scientific Factuality} measures the truthfulness and alignment of the generated responses with established scientific consensus, ensuring the dialogue remains accurate and grounded. We adopt LLM-as-a-Judge with strict and loose criteria: the strict criteria accept those that are complete in key points and consistent with the reference, while the loose mode allows those well-reasoned. \textbf{Audience Adaptation} assesses the capacity to dynamically tailor its complexity and tone to the user's specific background and expertise level. We introduce an audience adaptation rate (AAR), where the LLM judge determines the user's academic background based on the model's response style, and an instance that matches the actual label is considered successful. Finally, \textbf{Inquiry Efficiency} quantifies the system's ability to gather necessary information and guide the conversation toward resolution with minimal, highly informative conversational turns. It is measured by the average turns and the dialogue completion rate within limited turns. In our current experiments, we use GPT-5.6 Luna as our LLM judge, and human evaluations for consistency are presented in Appendix 8.


\subsection{Performance Analysis (Table~\ref{tab:dialogue_bench})}


\textbf{Scientific factuality remains the primary challenge for end-to-end speech dialogue systems.} Although recent approaches exhibit promising conversational capabilities, most of them still struggle to maintain scientific correctness during multi-turn interactions. Across offline end-to-end models, Qwen3-Omni achieves the best factual performance, reaching 94.84\% under the loose criterion and 50.99\% in strict settings, substantially outperforming previous open-source models by a large margin. In contrast, most systems obtain strict factuality scores below 20\%, indicating that while they often produce broadly relevant answers, they frequently omit essential scientific evidence or introduce subtle factual inconsistencies. The relatively large gap between loose and strict evaluation across all models further suggests that generating scientifically rigorous responses remains considerably more difficult than producing answers that are merely plausible.

\textbf{Audience adaptation is considerably more difficult than factual response generation.} All offline models achieve AAR around 45-54\%, suggesting limited capability in dynamically adjusting explanation depth according to users' expertise. The cascade system achieves the highest AAR (63.38\%), implying that strong LLMs retain superior instruction-following and controllability for style adaptation, whereas current speech models tend to generate with relatively fixed explanation styles regardless of the audience. Therefore, effectively incorporating content generation with audience-aware presentation remains an important direction.

\textbf{Inquiry efficiency strongly correlates with overall dialogue quality.} Qwen3-Omni requires only 5.85 turns on average to complete a dialogue while successfully finishing over 95\% of conversations within eight turns, indicating that it can efficiently identify missing information and progressively resolve users' questions. However, earlier speech-language models frequently require over ten turns, with many conversations failing to converge within the predefined dialogue budget. This behavior suggests that insufficient scientific understanding often leads to repetitive clarification requests or ineffective follow-up questions, reducing overall interaction efficiency.  The closed-source API, Qwen3.5-Omni-Flash, achieves the highest completion rate and the lowest average dialogue length, although its factuality remains below that of Qwen3-Omni. This observation indicates that optimizing dialogue strategy and scientific correctness are almost necessarily aligned objectives.


Overall, the results suggest that current models face three fundamental challenges in scientific dialogue: preserving factual consistency across multiple conversational turns, adapting explanations to users with diverse expertise, and efficiently acquiring missing information through goal-oriented interaction. Addressing these challenges will likely require tighter integration of scientific knowledge grounding, explicit user modeling, and dialogue policy optimization.
\section{Conclusion}


In this paper, we presented $S^3$-Bench designed for speech interaction models as scientific voice assistants. It comprises $S^3$-Knowledge for atomic question-answering and $S^3$-Dialogue, assessing multi-turn interactions. Our empirical evaluations yield several key insights. While upstream capabilities form a positively coupled chain with perception acting as a critical threshold, speech generation couples negatively due to a fundamental representation-pronunciation trade-off. Second, multi-turn dialogues expose persistent bottlenecks in maintaining strict factual consistency, dynamically adapting explanations to target audiences, and maintaining high inquiry efficiency. We hope $S^3$-Bench serves as a diagnostic benchmark to foster research toward tighter knowledge grounding, robust audio-text alignment, and user-adaptive modeling in spoken dialogue systems.

\clearpage
\bibliography{main}



\section*{Supplementary material}

# Appendix

## Contents

# 1 Appendix Introduction

Due to space constraints in the main text, we provide comprehensive details in the appendix. In the following content, Appendix 2 reviews related work, detailing recent advances in speech interactive models, speech evaluation frameworks, and scientific capability benchmarks. Appendix 3 elaborates on the dataset construction pipeline for the $S^3$-Knowledge and $S^3$-Dialogue benchmarks, while Appendix 4 reports complete statistical distributions for the final test set.

For implementation and reproducibility, Appendix 5 details model access, inference hyperparameter configurations, and representative model response examples. Appendix 6 introduces the fine-tuning variants of our speech recognition models for pronunciation evaluation, highlighting Qwen3-ASR-Terminology and Qwen3-ASR-Abbs. Appendix 7 presents additional experimental results omitted from the main text. Appendix 8 outlines our human evaluation protocol, designed to ensure speech quality and validate alignment with LLM-as-a-judge metrics. Finally, Appendix 9 lists the prompt templates used for automated capability evaluation and $S^3$-Dialogue system execution.

# 2 Related Works

## 2.1 Speech Interaction Models

Speech interaction models aim to directly comprehend vocal instructions and generate natural, expressive acoustic responses. Traditional cascade systems chain Automatic Speech Recognition (ASR), Large Language Models (LLMs), and Text-to-Speech (TTS) modules [1, 2, 3]. Despite their modularity, these systems suffer from accumulated latency, error propagation, and the loss of rich paralinguistic cues (e.g., pitch, tone, and emotional nuance) during the text bottleneck. To mitigate these limitations, early end-to-end (E2E) speech interaction models directly encoded speech queries without explicit text transcriptions to generate aligned text and speech outputs. These architectures predominantly diverge into two paradigms [4]:

- **Aligned Multimodal Models:** These systems utilize separate, modular acoustic decoders synced with a central LLM backbone to manage dual-modality outputs, represented by LLaMA-Omni [5], Freeze-Omni [6], VocalNet [7], and the Qwen-Omni series [8, 9]. In these models, speech generation is fully conditioned on the intermediate textual response produced by the backbone, making spoken outputs a faithful realization of the generated text. This design naturally aligns semantic content across modalities, and enables competitive speech interaction capabilities using comparatively limited speech training data.
- **Native Multimodal Models:** These models employ a unified, decoder-only Transformer architecture to jointly model and generate both textual and speech-related tokens in a single sequence, exemplified by Mini-Omni [10, 11], Moshi [12], GLM-4-Voice [13], and VITA-Audio [14]. By learning a shared representation space for multiple modalities, native architectures directly capture interactions between linguistic semantics and acoustic characteristics. Nevertheless, jointly modeling heterogeneous modalities substantially increases optimization complexity and typically requires large-scale multimodal pretraining data and carefully designed training strategies to achieve stable performance.

Building upon recent advancements, Speech LLMs and Omni-LLMs have progressively evolved toward specialized capabilities and more human-like, natural interaction paradigms. For instance, MiMo-Audio substantially scales up training data to achieve further performance gains [15]; models such as Covo-Audio and Fun-Audio-Chat introduce duplex architectures to decouple the speech perception and generation streams [16, 17]; and frameworks like VITA-Qinyu incorporate expressive features, including singing synthesis [18]. Concurrently, several high-performance models, such as Qwen3.5-Omni, are provisioned exclusively via closed-source APIs [19]. Since this study specifically focuses on multi-turn conversational scenarios, our scope is strictly confined to general-purpose speech interaction models.

## 2.2 Speech Interaction Evaluation

The evaluation of speech interaction models has transitioned from traditional, task-specific metrics to holistic, multi-turn, and vocal-centric conversational paradigms. We categorize existing

benchmarks into three progressive stages: task-oriented speech understanding, interactive speech generation, and conversational interactions in diverse scenarios.

Early evaluation frameworks primarily targeted text responses given vocal queries. AIR-Bench introduces a hierarchical evaluation framework comprising a multi-task foundation benchmark and an open-ended chat benchmark to systematically assess instruction-following capabilities across speech, environmental sound, and music [20]. AudioBench established a universal evaluation suite spanning 26 datasets across the understanding performance of voice, speech, and audio scene [21]. OpenAudioBench advanced this paradigm by assessing with knowledge-intensive question-answering (QA), reasoning queries, and daily chat instances [22]. However, these benchmarks only evaluate the model's text responses, ignoring the consistency and naturalness of the generated speech, thus cannot intuitively reflect the user's experience in speech interactions.

Subsequent works have increasingly focused on the consistency estimation of dual-modal responses alongside speech quality evaluation based on speech MOS, while further extending task scenarios including literary creation, rigorous multi-turn evaluation, and safety alignment. URO-Bench introduces a dual-track framework that facilitates multi-turn evaluation and cognitive reasoning, ensuring robust consistency estimation [23]. SOVA-Bench expands the evaluation paradigm to benchmark generative speech assistants across stages of recognition, understanding and generation, introducing paralinguistic-awareness assessment [24]. VocalBench establishes a holistic framework for semantic quality, acoustic performance, conversational abilities, and robustness, covering 14 user-oriented characters with diverse conversational settings [25].

Recent achievements have increasingly shifted toward conversational interactions in diverse, real-world scenarios. ParaS2SBench evaluates the naturalness of interactive pairs in content and speaking style using expressive and challenging queries [26]. It proposes a PolyTone training strategy preventing the style hallucination of audio LLM judging. Audio MultiChallenge stresses conversational models under heterogeneous acoustic conditions, measuring their robust performance across complex and noisy environments under natural multi-turn interactions [27]. VoiceAssistant-Eval complements these approaches by offering a system-level evaluation framework to assess AI assistants across listening, speaking, and viewing [28].

While the aforementioned benchmarks comprehensively encapsulate the vast majority of general speech interaction scenarios, a prominent gap remains: no existing evaluation framework is specifically tailored for specialized scientific dialogue settings. Given the profound domain expertise, structural complexity, and rigorous reasoning intrinsic to scientific discourse, it is highly imperative to construct a dedicated framework to evaluate the potential of models serving as advanced scientific assistants capable of speech interactions.

## 2.3 Scientific LLM Evaluation

The evaluation of scientific LLMs has progressively evolved from assessments of scientific knowledge to expert-level multimodal reasoning and, more recently, process-oriented evaluation of scientific discovery. Existing surveys categorize scientific LLM evaluation into three core dimensions: a) expert-level scientific knowledge comprehension and retrieval; b) scientific reasoning and problem solving; c) and multimodal scientific data understanding [29].

The first dimension focuses on evaluating whether models can acquire, retrieve, and apply scientific knowledge. General academic benchmarks such as MMLU and C-Eval cover scientific subjects ranging from secondary-school science to undergraduate-level physics, chemistry, and biology [30, 31]. As model performance on these benchmarks gradually improved, more challenging benchmarks were introduced to reduce the effectiveness of superficial pattern matching and elimination strategies. MMLU-Pro increases the number of candidate answers and incorporates more difficult distractors [32], while GPQA and SuperGPQA evaluate expert-authored questions that require graduate-level knowledge across biology, chemistry, physics, and other academic disciplines [33, 34]. These benchmarks provide standardized comparisons of scientific knowledge, but they predominantly rely on single-turn multiple-choice questions and final-answer accuracy, providing limited insight into intermediate reasoning processes or models' ability.

The second dimension evaluates scientific reasoning and problem-solving capabilities. Comprehensive benchmarks such as SciEval and SciKnowEval assess multiple levels of scientific competence, including knowledge recall, comprehension, reasoning, discernment, and application [35, 36]. Domain-specific benchmarks further incorporate the distinctive knowledge structures and reasoning requirements of individual scientific disciplines. For example, PHYBench evaluates advanced

physical reasoning and symbolic derivation, ChemBench assesses chemical knowledge and reasoning against human chemists [37, 38]. Such benchmarks increasingly adopt fine-grained, semantic, or domain-specific metrics rather than relying solely on exact-match accuracy. Nevertheless, most of them remain offline and task-isolated, making it difficult to assess interactive clarification, adaptive explanation, or error correction during scientific communication.

The third dimension concerns the understanding and reasoning of multimodal scientific data. ScienceQA combines textual questions with diagrams, images, and annotated explanations to evaluate multimodal scientific question answering [39]. MMMU and MMMU-Pro extend evaluation to expert-level reasoning over scientific figures, tables, charts, and interleaved visual-textual information [40, 41]. These benchmarks better reflect the multimodal nature of scientific knowledge, but they primarily consider visual, textual, symbolic, or structured inputs and rarely examine how specialized scientific information is conveyed and understood through spoken interaction.

Despite these advances, scientific LLM evaluation and speech-interaction evaluation have largely developed in isolation. Scientific benchmarks predominantly assess models through textual, visual, symbolic, or code-based inputs and outputs, leaving their ability to recognize specialized scientific expressions in speech and convey scientific knowledge through spoken responses largely unexplored. At the same time, existing speech interaction benchmarks mainly target general-domain conversations and provide limited coverage of technical terminology, spoken abbreviations, special expressions, rigorous scientific reasoning, and explanations adapted to users with different levels of expertise. Bridging these two evaluation paradigms is essential for assessing whether speech interaction models can function as reliable and effective scientific voice assistants.

# 3 Detailed Construction Pipeline

## 3.1 $S^3$-Knowledge

### 3.1.1 Public Scientific Benchmarks

We conduct an extensive survey and collection of publicly available scientific benchmarks. The selection criteria are as follows: (a) college-level or more challenging general scientific benchmarks with fine-grained disciplinary coverage, or evaluation datasets from specialized domains; (b) content that can plausibly occur in spoken conversations, excluding samples that require visual modalities or contain excessively complex structured text (e.g., SMILES strings and ultra-long gene sequences); (c) concise and well-defined information, excluding over-lengthy textual formats (e.g., electronic health records); and (d) contains challenging elements commonly encountered in spoken conversations, such as rare technical terminology, spoken norms of abbreviations, and the natural verbalization of symbolic special expressions. The retained benchmarks and corresponding information are summarized in Table 1.

| Benchmark | Domain | Task Type | Instances |
|---|---|---|---|
| SciKnowEval [36] | Chemistry, Life Sciences, Materials Science, Physics | MCQA | 581 |
| EarthSE [42] | Climate Science, Earth Science, Life Sciences | MCQA, Open-ended | 295 |
| GeneTuring [43] | Genomics | Open-ended | 216 |
| CS-Bench [44] | Computer Science | MCQA | 197 |
| MedMCQA [45] | Clinical Medicine | MCQA | 171 |
| BIOASQ [46] | Life Sciences | Open-ended | 95 |
| Lab-bench [47] | Life Sciences | MCQA | 88 |
| ChemBench [48] | Chemistry | MCQA, Open-ended | 79 |
| Genome-bench [49] | Genomics | MCQA | 50 |
| PubMedQA [50] | Clinical Medicine | MCQA | 48 |
| Astro-QA [51] | Astronomy | MCQA | 40 |
| ClimaQA [52] | Climate Science | MCQA, Open-ended | 52 |
| MedQA-USMLE [53] | Clinical Medicine | MCQA | 29 |
| MedXpertQA [54] | Clinical Medicine | MCQA | 26 |
| MaScQA [55] | Materials Science | MCQA | 7 |
| QCBench [56] | Chemistry | Open-ended | 4 |
| UGPhysics [57] | Physics | Open-ended | 2 |

Table 1: Data resources for $S^3$-Knowledge.

### 3.1.2 Text Pronunciation Calibration

Before speech synthesis, text queries undergo pronunciation calibration. Specifically, our inspection shows that the text-to-speech (TTS) model can correctly pronounce most terminology, as their pronunciations are closely correlated with their orthographic forms, and a large proportion consists of compound words with relatively straightforward syllabic pronunciations. In contrast, abbreviations and special expressions often follow standardized conventions established through long-term academic practice, resulting in pronunciations that differ substantially from their written forms. Accordingly, we conduct extensive manual annotation for these cases. To ensure a high standard of data quality and rigor, the annotation process was entirely executed by the author team. The annotator cohort possesses exceptional academic qualifications, with all members holding or currently pursuing a Bachelor's degree, and over half being graduate researchers (Master's or Ph.D. candidates). All annotators exhibit full professional proficiency and fluency in English, coupled with solid backgrounds in scientific methodologies. This unique combination of linguistic competence and specialized domain expertise guarantees precise comprehension and meticulous labeling of complex, technical scientific speech instances.

Specifically, by comparing them with authentic speech recordings from online video resources such as YouTube and Bilibili, we assign pronunciation-equivalent replacement forms for speech synthesis. For entries with limited video resources, we further infer pronunciations based on semantic information from professional academic websites. For example, abbreviations composed of consecutive consonants without an apparent lexical structure are pronounced letter by letter. A screenshot of the manual annotation interface is presented in Figure 1, which includes the original text question, the rule-based spoken form, the LLM-annotated academic full name, and direct links to online dictionaries and video resources for efficient verification.

### 3.1.3 Verification Process

To ensure the quality of the synthesized speech queries, we adopt a hybrid verification pipeline that combines automatic and manual verification. Specifically, the automatic verification stage employs dedicated automatic speech recognition (ASR) models for general speech, scientific terminology, and abbreviations, respectively, to transcribe the synthesized speech and compare the recognized content with the corresponding text segments. Any query containing mismatched segments is regenerated until no discrepancies remain.

The remaining samples after several rounds of re-synthesis are further manually verified using the annotation interface illustrated in Figure 2. We provide both the original text query and its spoken-form annotation, together with the corresponding synthesized speech segment and reference video with timestamp, enabling direct comparison between the synthesized pronunciation and authentic human speech. Whenever pronunciation inconsistencies are identified, the affected text is replaced with a pronunciation-equivalent form and re-synthesized until it passes manual verification.

### 3.1.4 Relevant Word Generation

To evaluate a model's perceptual capabilities, we introduce the Relevant Rate (RR), which measures whether a model response identifies the core entity, understands the queried concept, and connects them to relevant domain knowledge.

We extract candidate keywords for each question in $S^3$-Knowledge from four sources: the question, the answer, answer explanation, and reasoning chain generated by MiniMax-M3. Evidence from these sources is weighted by 5, 5, 3, and 2, respectively, to rank keyword candidates and select them.Then we filter out generic or low-information terms, retaining expressions that capture the core entities, target concepts, or knowledge anchors.

The extracted candidates are first classified into **core entities**, **target concepts**, and **knowledge anchors**. Core entities represent the main objects discussed in the question, while target concepts represent the properties, mechanisms, or concepts being queried. Knowledge anchors are subsequently derived from the supporting keywords, answer explanations, and reasoning chains to capture the domain-specific knowledge needed to connect the core entity and target concept.

All selected keywords must be traceable to the original source texts. Unsupported terms are removed. Samples with missing or low-quality annotations are further reviewed using Gemini-3.5-Flash. The resulting annotations are checked for empty fields, duplicate keywords, and unsupported

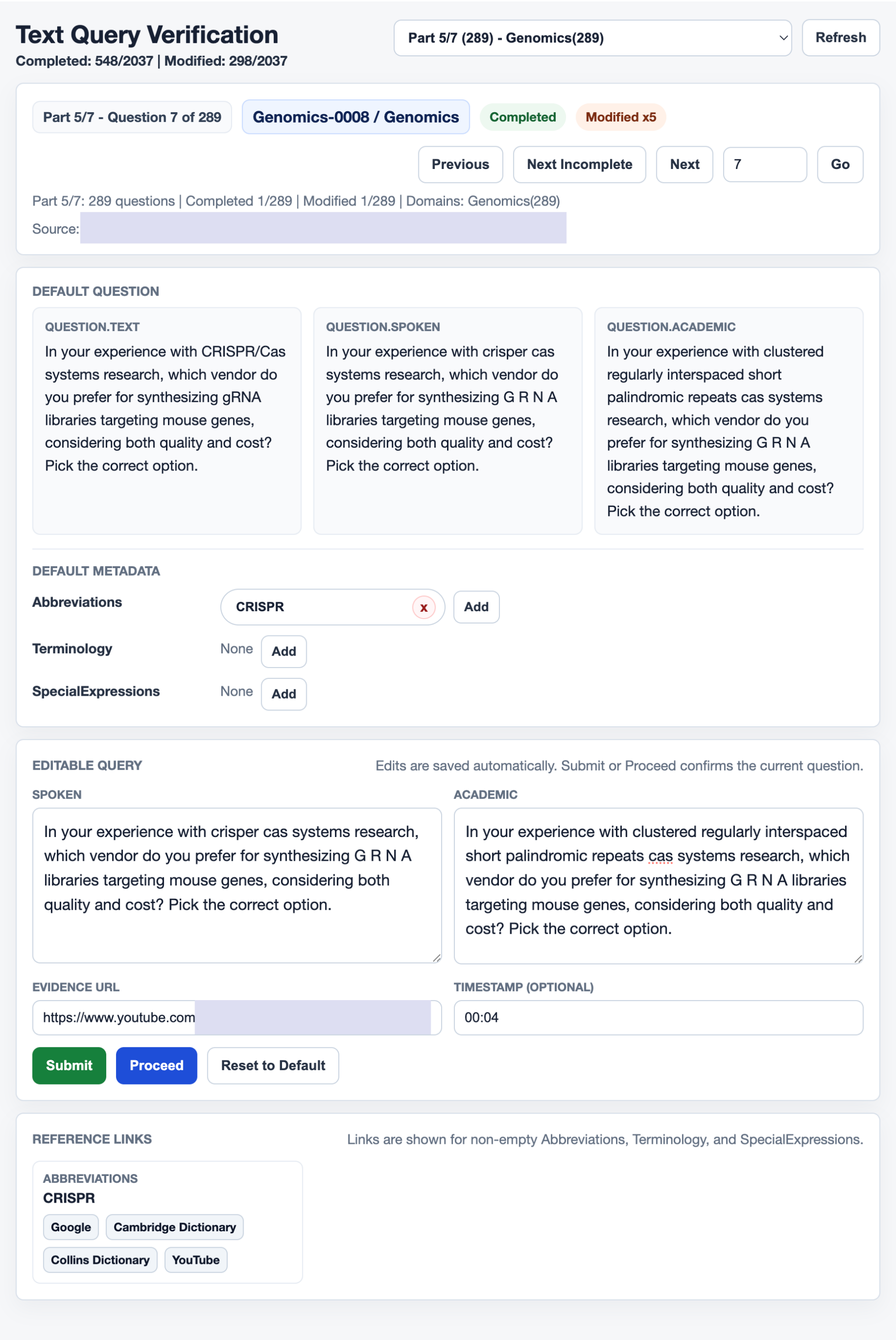


Figure 1: The screenshot of the annotation process for query pronunciations.

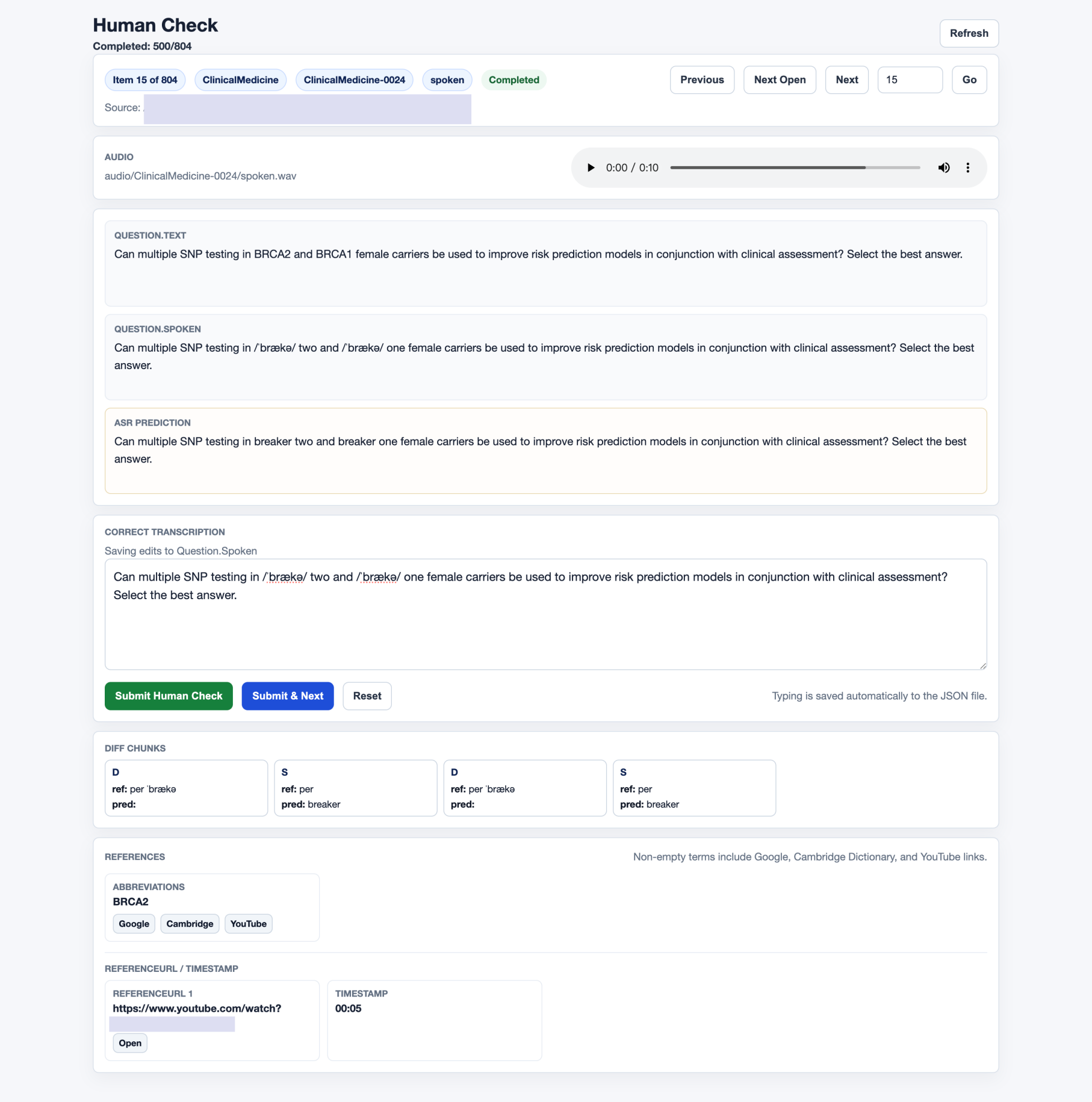


Figure 2: The screenshot of the manual verification process after speech synthesis.

terms. During the perception evaluation phase, the model is provided only with the speech query, without any additional text prompts. If the model's text response in this open-ended QA setting contains any item in the keyword list, it is considered that the model has successfully perceived the key semantics.

## 3.2 $S^3$-Dialogue

### 3.2.1 Topic Selection

The goal of topic selection is to identify scientific topics from Darwin-Science that are suitable for constructing $S^3$-Dialogue, rather than simply extracting isolated keywords or complete paper titles [58].

We organize the candidate pool into ten scientific domains: computer science, physics, chemistry, materials science, clinical medicine,life sciences, genomics, earth science, climate science, and astronomy.Within each domain, we prioritize specific scientific concepts, models, methods, mechanisms, experimental techniques, evaluation metrics, and application scenarios.

The topic selection process consists of four main steps:

1. Candidate phrases are extracted from the Darwin-Science corpus, while complete paper titles are avoided.

2. Short and information-dense topics, typically containing approximately two to six English words, are prioritized because they are more suitable for multi-turn scientific dialogue.

3. The remaining candidates are further selected according to domain coverage, topic specificity, mechanism interpretability, application scope, and their potential to support subsequent dialogue generation.

The final candidate set is additionally balanced across scientific domains.The current working version contains 213 topics, with each domain containing at least 20 topics. The domain distribution is shown in Table 2.

| **Scientific Domain** | **Number of Topics** |
|---|---|
| Computer Science | 32 |
| Clinical Medicine | 21 |
| Physics | 20 |
| Chemistry | 20 |
| Materials Science | 20 |
| Life Sciences | 20 |
| Genomics | 20 |
| Earth Science | 20 |
| Climate Science | 20 |
| Astronomy | 20 |
| **Total** | **213** |

Table 2: Distribution of the selected scientific topics across domains.

### 3.2.2 Dialogue Construction

The goal of dialogue construction is to transform each scientific topic into an evaluable multi-turn dialogue sample. Each sample is centered on a single topic and contains a user profile, an initial question, several follow-up questions, and corresponding reference answers.

The user profile controls both the difficulty and interaction style of the dialogue. The difficulty level is selected from *Beginner*, *Undergraduate*, *Graduate*, and *Expert*, while the user character is selected from *curious*, *patient*, *anxious*,*overconfident*, and *practical.* These attributes guide the depth, terminology, assumptions, and conversational style of the generated questions and answers.

The initial question and follow-up questions jointly cover five complementary aspects of the topic:

- **Concept.** This category asks what the topic is and establishes its basic definition, central intuition, and conceptual scope. It tests whether the model can explain the core idea of a scientific topic, identify its main components or variables, and distinguish it from closely related concepts. The expected depth increases with the user profile: simpler dialogues emphasize intuitive explanations and concrete examples, while more advanced dialogues require clearer conceptual boundaries, assumptions, operational definitions, and conditions under which the concept is valid.
- **Mechanism.** This category asks how the topic works, focusing on internal processes, procedural steps, and mathematical or experimental principles. It is intended to test mechanistic understanding rather than superficial repetition of definitions.
- **Milestone.** This category asks why the topic became important, referring to influential papers, experimental techniques, databases, observations, or engineering platforms. When a specific year or single historical source cannot be established reliably, the question instead focuses on the technological or practical developments that increased the importance of the topic.
- **Characteristics.** This category corresponds to the strengths and limitations of the topic.It examines advantages, weaknesses, applicable conditions, failure cases, and common misconceptions. This helps prevent unsupported generalization beyond the setting in which a method or finding has been validated.
- **Application.** This category examines how the topic is used in real research, engineering, or evaluation scenarios. It also considers conditions under which the topic may fail.

Overall, each $S^3$-Dialogue sample evaluates not only the concept,but mechanism, importance, characteristics, and applications of a scientific topic as well.

# 4 Benchmark Statistics

The data distribution of the $S^3$-Knowledge benchmark is summarized in Table 3. It covers 10 major scientific domains and is constructed from 17 publicly available benchmark datasets, comprising a total of 1,980 test samples. The average query duration is 12.74 seconds, while the longest samples exceed 20 seconds. Several domains incorporate data from multiple benchmark sources. For example, the Life Sciences domain includes samples collected from four different benchmarks. Conversely, a single benchmark may contribute to multiple domains because it covers diverse scientific topics. For instance, EarthSE provides samples spanning the Life Sciences, Earth Sciences, and Climate Science domains [42].

The domain distribution of the $S^3$-Dialogue benchmark and the average number of words across the five progressive topic aspects are summarized in Table 4. Most domains contain 20 dialogue topics, while Computer Science includes the largest number, with a total of 32 dialogues. In addition, the lengths of user queries remain largely consistent across the first four aspects, whereas the final Application aspect generally contains longer queries. All dialogue topics are listed in Table 5.

# 5 Evaluated Models

## 5.1 Model Overview

The evaluated models in $S^3$-Bench are listed in Table 6. Our benchmark currently includes six Speech LLMs, two Omni-LLMs, and one proprietary API. Most multimodal LLMs adopt a native architecture, where speech and text are jointly modeled by a shared backbone. During decoding, speech tokens are either incorporated into the shared decoding head or generated in parallel using separate decoding heads. For the cascaded baseline, we employ Whisper-Large-v3 as the ASR model, as it demonstrates strong performance in recognizing scientific terminology and abbreviations (Table 1 in the main text). For speech generation, we use the custom voice mode of Qwen3-TTS with the default Ryan voice.

| Domain | Dataset | Instances | Avg Duration (s) | Avg Num of Words |
|---|---|---|---|---|
| Physics | SciKnowEval | 581 | 13.11 | 27.97 |
| | UGPhysics | 2 | 17.08 | 37.50 |
| Computer Science | CS-Bench | 197 | 8.17 | 19.62 |
| Genomics | GeneTuring | 216 | 6.32 | 9.30 |
| | Genome-Bench | 50 | 15.31 | 34.62 |
| Life Sciences | SciKnowEval | 66 | 11.34 | 23.26 |
| | EarthSE | 18 | 19.82 | 45.89 |
| | BioASQ | 95 | 5.19 | 8.21 |
| | Lab-Bench | 88 | 10.80 | 22.14 |
| Clinical Medicine | MedMCQA | 171 | 18.20 | 33.71 |
| | PubMedQA | 48 | 10.05 | 18.73 |
| | MedQA-USMLE | 29 | 21.56 | 49.10 |
| | MedXpertQA | 26 | 18.73 | 40.19 |
| Earth Science | EarthSE | 122 | 18.46 | 43.85 |
| Climate Science | EarthSE | 155 | 17.88 | 44.06 |
| | ClimaQA | 52 | 10.78 | 24.94 |
| Astronomy | Astro-QA | 40 | 12.50 | 29.20 |
| Material Science | SciKnowEval | 246 | 14.45 | 30.19 |
| | MaScQA | 7 | 14.54 | 34.00 |
| Chemistry | ChemBench | 79 | 13.63 | 24.94 |
| | SciKnowEval | 36 | 20.29 | 35.19 |
| | QCBench | 4 | 17.66 | 34.00 |
| Total | 17 Sources | 1980 | 12.74 | 27.03 |

Table 3: $S^3$-Knowledge scientific area distribution.

| Domain | Instances | Concept | Mechanism | Milestone | Characteristics | Application |
|---|---|---|---|---|---|---|
| Physics | 20 | 14.40 | 15.15 | 11.50 | 15.10 | 15.05 |
| Computer Science | 32 | 15.78 | 15.16 | 14.38 | 14.94 | 18.50 |
| Genomics | 20 | 14.90 | 14.65 | 12.40 | 15.40 | 18.55 |
| Life Sciences | 20 | 14.80 | 14.45 | 14.80 | 14.75 | 18.55 |
| Clinical Medicine | 21 | 15.38 | 15.86 | 13.90 | 14.33 | 16.52 |
| Earth Science | 20 | 14.20 | 13.85 | 14.05 | 13.70 | 16.20 |
| Climate Science | 20 | 15.45 | 15.00 | 12.85 | 14.05 | 15.60 |
| Astronomy | 20 | 14.70 | 15.45 | 13.55 | 15.25 | 17.10 |
| Material Science | 20 | 14.80 | 15.80 | 12.65 | 14.30 | 16.90 |
| Chemistry | 20 | 15.00 | 13.35 | 13.90 | 14.90 | 17.10 |
| Total | 213 | 14.99 | 14.89 | 13.46 | 14.69 | 17.09 |

Table 4: $S^3$-Dialogue scientific area distribution and average word num in each aspect.

| Domain | Topics |
|---|---|
| **Physics** | Lattice QCD simulations, Euclidean path integrals, Wilson loops, QCD sign problem, Quantum entanglement entropy, Noether symmetry currents, Lagrangian versus Hamiltonian mechanics, Higgs mechanism, Dirac equation spinors, Casimir effect, Landau levels, Magnetic reconnection, Blackbody radiation, Boltzmann distribution, Fermi surface, Josephson junctions, Spin-orbit coupling, Anomalous magnetic moment, Perturbation theory breakdown, MHD turbulence. |
| **Computer Science** | hybrid feature selection for MLP intrusion detection, emotional empathy in small language models, empathy evaluation for dialogue agents, knowledge distillation for small language models, retrieval-augmented generation hallucination evaluation, prompt injection defense, tool-use evaluation for agents, vector index recall, retrieval chunking strategies, contrastive learning for embeddings, graph neural network message passing, Transformer positional encodings, rotary position embeddings, attention head pruning, mixture-of-experts routing, speculative decoding, KV cache compression, Out-of-distribution detection, Adversarial examples in classifiers, Differential privacy SGD, Federated averaging, Dijkstra priority queue implementation, Nucleus sampling, Batch normalization, Layer normalization, Label smoothing cross-entropy, Focal loss for class imbalance, Active learning uncertainty sampling, Semi-supervised pseudo-labeling, Data leakage detection, Stratified train-test split, UNSW-NB15 intrusion detection. |
| **Genomics** | Short-read sequencing, Base calling accuracy, Read alignment algorithms, De novo genome assembly, Indel calling, Genotype imputation, Polygenic risk scores, Expression quantitative trait loci, RNA-seq differential expression, Spatial transcriptomics, ChIP-seq peak calling, CUT&Tag, DNA methylation arrays, Bisulfite sequencing, DPWG dosing guidelines, gnomAD allele frequency, Rare variant burden tests, Trio sequencing, Whole-genome sequencing data, SNP annotation. |
| **Life Sciences** | Ras-MAPK signaling pathway, TCR signaling in NKT cells, PLZF expression in NKT development, GPCR signal transduction, Kinase phosphorylation cascades, Protein function annotation, Autophagy flux, Apoptosis versus necrosis, Antigen presentation, Microbiome-host interactions, Epithelial barrier function, Osmosis, Membrane potential, Ion channel gating, Retinal pigment epithelial cells, Hormone receptor signaling, Stem cell differentiation, Positive selection in thymus, Clonal selection theory, Protein misfolding. |
| **Clinical Medicine** | Diagnostic workup, Targeted antifungal therapy, Echinocandin resistance, Amphotericin B nephrotoxicity, Therapeutic drug monitoring, Pharmacogenomic dosing, Drug-drug interactions, Renal function tests, Diagnostic sensitivity and specificity, Likelihood ratios, Positive predictive value, QT interval prolongation, Pulse oximetry limitations, Blood pressure regulation, Antibiotic stewardship, Deep vein thrombosis risk stratification, Pulmonary embolism workup, Autoimmune disease flares, Rheumatoid arthritis biologics, Inflammatory bowel disease biomarkers, Melanoma targeted therapy. |
| **Earth Science** | Magma viscosity, Eruption forecasting, Plate tectonics, Fault mechanics, Tsunami generation, Magnetic reversal timescale, Crustal deformation, Rock cycle, Sedimentary rocks, Depositional facies, Stratigraphic correlation, Fossil record bias, Compression fossils, Marine sediments, Lacustrine sediments, Mineral formation, Soil moisture sensors, Groundwater aquifers, Volcanic activity, Carbonate rocks. |
| **Climate Science** | Ocean acidification under climate change, Climate model predictions, Carbon budget, Radiative forcing, Climate sensitivity, Equilibrium climate sensitivity, Aerosol-cloud interactions, Ice sheet tipping points, Urban heat island, Heatwave attribution, Extreme precipitation attribution, Monsoon variability, Stratospheric ozone recovery, Paleoclimate reconstruction, Sea level rise projections, Storm surge risk, Soil moisture drought, Climate adaptation planning, Transient climate response, Atlantic meridional overturning circulation. |
| **Astronomy** | Dark matter detection, QCD axion dark matter, WIMP direct detection, Cosmic microwave background anisotropy, Cosmic inflation, Cepheid distance ladder, Active galactic nucleus feedback, Relativistic jets, Gravitational lensing, Exoplanet transit method, Tidal locking, White dwarf cooling, Pulsar timing arrays, Fast radio bursts, Solar wind, Coronal mass ejections, Magnetospheres, Protoplanetary disks, Type Ia supernova cosmology, Radial velocity method. |
| **Material Science** | NaYF4 host matrix, Surface quenching in UCNPs, Grain boundary passivation, Thin-film annealing, XRD peak broadening, Scherrer equation, TEM lattice fringes, AFM surface roughness, Electrical conductivity stability, Surface passivation, Strain engineering, Doped semiconductors, Solid electrolyte interface, Lithium dendrite suppression, Fatigue in alloys, Ferroelectric domain switching, Thermoelectric figure of merit, Phonon scattering, Thermal conductivity measurement, Up-conversion nanoparticles. |
| **Chemistry** | Henderson-Hasselbalch equation, Thin-layer chromatography Rf, UV-Vis absorption spectra, Fluorescence quenching, Reaction rate law, Arrhenius activation energy, E1 versus E2 elimination, Nucleophilic aromatic substitution, Suzuki coupling, Ligand field splitting, Crystal field theory, Coordination geometry, Chelation effect, Acid-base titration curves, Buffer capacity, Beer-Lambert law, pH and pKa, SN1 versus SN2 reactions, HPLC retention time, FTIR functional group assignment. |

Table 5: Topics across domains in $S^3$-Dialogue.

| Model | Model Card | Base LLM | Modality | Decoding Type |
|---|---|---|---|---|
| VocalNet [7] | VocalNet-Qwen3-8B | Qwen3-8B | $T \cup S$ | Aligned |
| VITA-Audio [59] | VITA-Audio-Plus-Vanilla | Qwen2.5-7B | $T \cup S$ | Native |
| Kimi-Audio [60] | Kimi-Audio-7B-Instruct | Qwen2.5-7B | $T \cup S$ | Native |
| MiMo-Audio [15] | MiMo-Audio-7B-Instruct | MiMo-7B-Base | $T \cup S$ | Native |
| Step-Audio-2-mini [61] | Step-Audio-2-mini | Qwen2.5-7B | $T \cup S$ | Native |
| Fun-Audio-Chat [17] | Fun-Audio-Chat-8B | Qwen3-VL-8B | $T \cup S$ | Native |
| Qwen2.5-Omni [8] | Qwen2.5-Omni-7B | Qwen2.5 | $T \cup S \cup V$ | Aligned |
| Qwen3-Omni [9] | Qwen3-Omni-30B-A3B-Instruct | Qwen3 | $T \cup S \cup V$ | Aligned |
| Qwen3.5-Omni-Flash [62] | qwen3.5-omni-flash-2026-03-15 | - | $T \cup S \cup V$ | API |
| Cascade | whisper-large-v3, Qwen3-8B, Qwen3-TTS-12Hz-1.7B-Base | Qwen3-8B | $T \cup S$ | Cascade |

Table 6: Evaluated models of $S^3$-Bench. For modality, $T$, $S$, $V$ refer to text, speech and vision respectively.

## 5.2 Model Configuration

During inference, we use the default generation parameters and voice settings whenever possible. For models that do not provide a built-in voice, we use a clean English male utterance from the Common Voice dataset as the speech prompt. All experiments are conducted on NVIDIA L20 GPU(s), except for Kimi-Audio, which is evaluated on an NVIDIA A100 GPU due to its relatively high GPU memory requirements. The voice settings and dialogue history modeling strategies used for each model are summarized in Table 7. For the ASR task, we use the prompts listed in Table 8. For the multiple-choice question answering (MCQA) task, we additionally provide the textual prompt in the following format: "Options:\nA. XXX\nB. XXX\n...".

| Model | Timbre | User History | Assistant History |
|---|---|---|---|
| VocalNet | common_voice_en_2586258.wav | Whisper ASR text | Text reply |
| VITA-Audio | default | Audio (path list resolved by placeholders) | Text |
| Kimi-Audio | default | None (single-turn) | None (single-turn) |
| MiMo-Audio | prompt_speech_en_m.wav | Audio (path) | Audio + text (`{text, audio_path}`) |
| Step-Audio-2-mini | default | Audio (`{type:audio}`) | Speech tokens (server `tts_content`) |
| Fun-Audio-Chat | common_voice_en_2586258.wav | Audio (accumulated 16 kHz waveforms) | Text |
| Qwen2.5-Omni | default | Audio (re-read `.wav` path each turn) | Text |
| Qwen3-Omni | default Ethan | Audio (re-read `.wav` path each turn) | Text |
| Qwen3.5-Omni-Flash | default Cherry | Audio (re-encoded base64 `input_audio` each turn) | Text |
| Cascade | common_voice_en_2586258.wav | Whisper ASR text | Text |

Table 7: Model configuration of the evaluated models of $S^3$-Bench.

| Model | ASR Prompt |
|---|---|
| VITA-Audio | Convert the speech to text. |
| Kimi-Audio | Please transcribe the following audio: |
| MiMo-Audio | Please transcribe the speech into text. |
| Step-Audio-2-mini | default Mandarin prompt |
| Fun-Audio-Chat | Please help me transcribe the audio. |
| Qwen2.5-Omni | Transcribe the audio into text verbatim. |
| Qwen3-Omni | Transcribe the audio into text. |
| Qwen3.5-Omni-Flash | |

Table 8: Speech recognition prompts for evaluated models.

## 5.3 Model Response Example

### 5.3.1 $S^3$-Knowledge

Model response examples on the $S^3$-Knowledge benchmark are presented in Table 10. Due to space limitations, we omit some intermediate outputs. Some models tend to generate concise responses, resulting in relatively clear speech responses. In contrast, models represented by Qwen3-Omni employ a more capable text LLM as the thinker; however, the highly structured textual outputs introduce pronunciation ambiguities for the talker, leading to numerous substitution errors as well as truncation and distortion when synthesizing responses containing long symbolic expressions. Additionally, we observe that certain models feature speech-adapted backbones, represented by MiMo-Audio, which yield relatively clear and natural speech output despite producing less readable text responses.

| Model | Text Response | Speech Response |
|---|---|---|
| VocalNet | To find the number of structural isomers for C8H18, we need to consider the different ways to arrange the carbon atoms. C8H18 is an alkane, and the number of structural isomers for alkanes can be determined by the number of ways to branch the carbon chain. For C8H18, there are 18 structural isomers. | To find the number of structural isomers for C(CH)H, we need to consider the different ways to arrange the carbon atoms. C(CH)H is an alkane, and the number of structural isomers for alkanes can be determined by the number of ways to branch the carbon chain. For C(CH)H, there are 18 structural isomers. |
| VITA-Audio | For C8H18, which is octane, there are 7 structural isomers. These are the straight-chain and branched-chain forms without considering the different spatial arrangements. | For C8H18, which is octane, there are seven structural isomers. These are the straight chain and the branched chain forms, without considering the different spatial arrangements. |
| Kimi-Audio | For C8H18, there are 10 structural isomers. That's when you have the same atoms but arranged in different ways. | For CBHCAT, there are ten structural isomers. That's when you have the same atoms but arranged in different ways. |
| MiMo-Audio | To determine the number of structural isomers for C eight H eighteen, ... For eight carbon atoms, we can list all the unique structures. These are: one, n-octane, which is a straight chain. Two, two-methylheptane. Three, three-methylheptane. Four, four-methylheptane. Five, two-two-dimethylhexane. Six, two-three-dimethylhexane. Seven, two-four-dimethylhexane. Eight, three-three-dimethylhexane. Nine, three-four-dimethylhexane. And ten, two-two-three-trimethylpentane.\n\nSo, in total, there are 10 structural isomers for C eight H eighteen, not including stereoisomers like geometric or optical isomers. | To determine the number of structural isomers for C8H18, ... For eight carbon atoms, we can list all the unique structures. These are 1 and octane, which is a straight chain; 2, 2-methylheptane; 3, 3-methylheptane; 4, 4-methylheptane; 5, 2, 2-dimethylheptane; 6, 2, 3-dimethylheptane; 7, 2, 4-dimethylheptane; 3, 3-methylhept |
| Step-Audio-2-mini | There are 10 structural isomers excluding stereochemistry for C8H18. | There are one ten structural isomers, excluding stereochemistry, for C8H8. |
| Fun-Audio-Chat | There are 18 structural isomers for C8H18, excluding stereochemistry. | There are seventeen structural isomers for C8H18, excluding stereochemistry. |
| Qwen2.5-Omni | Well, for $C_8H_{18}$, there are 3 structural isomers. One is n - octane, which is a straight - chain alkane. Another is 2 - methylheptane, and the third is 3 - methylheptane. So, that's the answer. If you have any other questions about chemical isomers or anything else related, feel free to ask. | Well, for C8H18, there are three structural isomers. One is n-octane, which is a straight chain alkane. Another is 2-methylheptane, and the third is 3-methylheptane. So that's the answer. If you have any other questions about chemical isomers or anything else related, feel free to ask. |
| Qwen3-Omni | To determine the number of **structural isomers** (also called **constitutional isomers**) for the molecular formula $**C_8H_{18}**$, excluding stereochemistry, we need to find all the different ways to arrange 8 carbon atoms and 18 hydrogen atoms such that each carbon has four bonds and each hydrogen has one bond.\n\n This formula corresponds to an **alkane** because it fits the general formula for alkanes: \n**$C_nH_{2n+2}$** ...\n - 2,3-dimethylhexane \n - 2,4-dimethylhexane \n - 2,5-dimethylhexane \n - 3,3-dimethylhexane | To determine the number of structural isomers, also called constitutional isomers, for the molecular formula C8H18I, excluding stereochemistry, we need to find all the different ways to arrange 8 carbon atoms and 18 hydrogen atoms such that each carbon has 4 bonds and each hydrogen has 1 bond. This formula corresponds to an alkane because it fits the general formula for alkanes, CH2sVH2, ... |
| Qwen3.5-Omni-Flash | There are 18 structural isomers for C8H18, excluding stereochemistry. | There are 18 structural isomers for C8H18, excluding stereochemistry. |
| Cascade | The number of structural isomers (excluding stereochemistry) for $C_8H_{18}$ is **18**. | The number of structural isomers (excluding stereochemistry) for C dash C fifth C one O is eighteen. |

Table 10: Model response example in $S^3$-Knowledge. Question: How many structural isomers (excluding stereochemistry) are there for C8H18?

### 5.3.2 $S^3$-Dialogue

Model response examples in $S^3$-Dialogue are presented in Tables 12 and 14. Due to space constraints, we present composite dialogues by stitching responses across different evaluation aspects from multiple models under the same topic. That is, these examples are not complete, continuous dialogues, while reflecting the model's responsiveness and style.

| Question | Turn | Model | Response |
|---|---|---|---|
| I'm curious about the precise operational definition of vector index recall in high-dimensional spaces, and specifically, what factual claim about this metric actually requires rigorous evidence to validate? | 1-1 | VocalNet | Vector index recall in high-dimensional spaces refers to the ability of an indexing method to efficiently retrieve vectors that are closest to a given query vector. This metric is crucial for applications like search engines, recommendation systems, and machine learning. The operational definition involves measuring how often the top-k nearest neighbors to a query vector are correctly identified by the index. The factual claim that requires rigorous evidence to validate is the accuracy and efficiency of the indexing method. Specifically, it involves proving that the method can consistently retrieve the nearest neighbors within a certain time frame and with a high degree of accuracy, even as the dimensionality and size of the dataset increase. This validation typically involves extensive testing and benchmarking against other indexing methods. |
| Before I can fully trust that empirical recall number, what foundational assumptions actually make vector index recall theoretically valid, and how exactly would I design an experiment to test if they hold in practice? | 2-1 | VITA-Audio | To design an experiment to test the theoretical validity of foundational assumptions for vector index recall, you should start by identifying the key assumptions. These might include the distribution of data, the accuracy of the similarity metric used, and the efficiency of the index structure. Next, gather a representative sample of your data and use it to create a baseline recall measurement. Then, systematically vary each assumption one at a time while keeping other factors constant. For example, change the data distribution, use different similarity metrics, or modify the index structure. Measure the recall in each scenario and compare it to the baseline. By analyzing these results, you can determine if the assumptions hold in practice. If the recall significantly deviates from the expected values, it indicates that the assumptions may not be valid under those conditions. |
| You mentioned that the data must be distributed in a way that allows efficient indexing, but what specific topological properties of the vector space, such as intrinsic dimensionality or local manifold structure, mechanically dictate the upper bound of recall for a graph-based index? | 3-1 | Kimi-Audio | The upper bound of recall for a graph-based index is influenced by the intrinsic dimensionality and local manifold structure of the data. Think of it like this: if your data is spread out in a high-dimensional space, it's harder to find all the relevant points because they're far apart. But if your data is clustered in a lower-dimensional space, you can find the relevant points more easily. The local manifold structure is like the shape of the data – if it's all tangled up, it's harder to find the right points. So, the more spread out or tangled your data is, the lower the recall you can expect from a graph-based index. |

| Question | Turn | Model | Response |
|---|---|---|---|
| We've covered the theoretical bounds, but I'm curious about the historical catalyst for this metric. Was there a specific benchmark like ANN-benchmarks, or a highly influential deployed system like FAISS, that really made optimizing vector index recall a critical priority in the field? | 3-2 | Kimi-Audio | The focus on optimizing vector index recall can be traced back to the rise of large-scale machine learning and the need for efficient similarity search. The development of ANN benchmarks and the success of systems like FAISS have indeed played a significant role in making this a priority. These tools and frameworks have made it easier to experiment and optimize, driving the field forward. |
| You mentioned ANN-benchmarks and FAISS as the main catalysts, but I'm curious about what specifically about the ANN-benchmarks framework changed our evaluation paradigm. Did the standardization of the recall versus queries-per-second trade-off curve force the community to treat recall as a tunable hyperparameter rather than just a static evaluation metric? | 3-3 | Kimi-Audio | Yeah, ANN Benchmarks really changed the game. It's not just about the recall vs queries per second trade-off curve, but also the way it standardizes the evaluation process. It forces the community to treat recall as a tunable hyperparameter, not just a static evaluation metric. This shift in perspective has led to a more nuanced understanding of the trade-offs involved in ANN systems. |
| Since FAISS cemented recall@k as the standard, I'm really curious about how the metric's own properties affect our methodology. When designing experiments to evaluate these systems, what specific strengths and inherent limitations of vector index recall actually dictate how we structure the study and interpret the results? | 4-1 | MiMo-Audio | That's a crucial consideration. The properties of recall at k have significant implications for both how you design your experiment and how you interpret the results. Let's start with its strengths. First, it has a very clear alignment with real-world needs, since many applications like recommendation systems or image retrieval prioritize finding the very top results. Second, it's quantitatively robust and easy to compare across different systems. Third, it directly measures the index's quality. And finally, it's a flexible metric that works well with other metrics like precision or NDCG. Now, for the limitations and their experimental implications. A major one is that recall at k ignores irrelevant results; it only cares about the top k. To counter this, your experiments should also measure metrics like precision at k or the F-beta score. Another key limitation is its sensitivity to the choice of k. To handle this, you should report results at multiple k values and analyze how the performance changes. The metric also assumes a fixed definition of 'relevance', so you must clearly define this in your study, perhaps through user studies. It also assumes a fixed query, so you should test it on diverse queries. Lastly, recall at k doesn't measure efficiency. So, your experiments must always report the query time and memory footprint alongside the recall scores. In summary, when you're designing an experiment, you need to clearly define your ground truth, choose your k values strategically, measure complementary metrics, and report both recall and efficiency. |

| Question | Turn | Model | Response |
|---|---|---|---|
| "Pivoting from these theoretical mechanics to concrete application, where exactly is vector index recall integrated into downstream models, benchmarks, or evaluation pipelines, and what specific failure should we be checking for to catch silent degradations in those environments? | 5-1 | Step-Audio-2-mini | Vector Index Recall is a key component of downstream models, benchmarks, and evaluation pipelines in the vector search community. It's often used as one of several metrics to evaluate the performance of the entire search system, alongside precision, mean average rank, and mean average precision. To catch silent degradations, we should check for: 1. Drift in neighbor quality - silent changes in the index's quantization scheme can lead to performance degradation without obvious signs. 2. Robustness to input noise - Internal noise in the index or downstream models can mask gradual performance degradation. 3. Generalizability - Check how the index's performance changes when deployed in diverse real-world scenarios. 4. Latency and throughput - Some algorithms may perform well in terms of recall but suffer from poor latency or scalability, which affects real-world usage. |
| Computing exact ground-truth recall in a live pipeline is computationally prohibitive, which makes continuous monitoring difficult. Given this constraint, what specific proxy signals or lightweight telemetry do engineers actually instrument to catch silent degradations like quantization drift before they impact the end user? | 5-2 | Step-Audio-2-mini | In practice, engineers often instrument the system with proxies that can detect subtle changes before major degradations occur. These include: 1. Latency monitoring - track response times to spot potential degradation early. 2. Precision metrics - measure how well the search results align with user expectations. 3. Diversity preservation - ensure the model maintains a balanced representation of data. 4. User feedback - Aggregate and analyze user reports to identify issues. 5. Model health checks - Continuously check model health through internal metrics to catch potential problems. |

Table 12: Model response example in $S^3$-Dialogue. Topic: vector index recall.

| Question | Turn | Model | Response |
|---|---|---|---|
| I need to run some rare variant burden tests for my project, so could you explain what they actually do in practice and what the key trade-off is that I should keep in mind? | 1-1 | Fun-Audio-Chat | Sure. Rare variant burden tests aggregate rare variants in a gene or region to see if they collectively affect a trait, often using a weighted or uniform score and a gene-level test like SKAT or VEGAS. The key trade-off is power: they gain power for very rare variants (MAF <0.01) but lose it for common ones, and they're sensitive to missense vs. loss-of-function bias and population stratification. |
| When I'm designing my assay, prepping real sequencing data, or doing genomic interpretation, what is the first mechanism of the rare variant burden test that I should actually check to make sure it's set up right? | 2-1 | Qwen2.5-Omni | Well, when you're setting up a rare variant burden test, the first thing you should check is the definition of r̈arev̈ariants. Different tools might have different thresholds for what they consider rare. For example, some might define it as variants with a frequency of less than 1% in a certain population. You also need to make sure that the test is correctly grouping the variants. It should be grouping all the rare variants in the right gene or region. And don't forget to check the statistical model. It should be appropriate for the type of data you have.If you want to know more about any of these aspects or have other questions related to this, feel free to ask. |

| Question | Turn | Model | Response |
|---|---|---|---|
| Since I'm trying to connect this to real-world applications, which specific study or result first showed the actual need for rare variant burden tests and made them visible as a scientific problem? | 3-1 | Qwen3-Omni | The need for rare variant burden tests really became clear in the early 2010s, especially with studies on severe developmental disorders and rare diseases. One of the first big examples was research into autism spectrum disorder and intellectual disability. Before that, genome-wide association studies (GWAS) were great for finding common variants linked to diseases, but they missed rare ones. When researchers started sequencing the exomes of affected individuals and their parents, they found that many people with these conditions had rare, damaging mutations in certain genes. But because each mutation was so rare, you couldn't see a signal by looking at them one by one. That's when burden tests came in—they pooled these rare variants and showed that people with the disorder had a higher burden of them in specific genes. This was a turning point, showing that rare variants collectively could play a big role in disease, and that traditional methods weren't enough. |
| Now that I'm ready to apply this to my own sequencing data, what main practical trade-off should I check before I fully rely on a rare variant burden test? | 4-1 | Cascade | Before fully relying on a rare variant burden test, the main practical trade-off you should check is the **balance between sensitivity and specificity**. Burden tests are powerful for detecting genes with multiple rare variants that collectively contribute to a trait, but they can be less effective if the variants are not clustered or if the effect sizes are too small. Additionally, they may miss variants that are individually significant but rare. You should also consider whether your study has enough statistical power—this depends on sample size, variant frequency, and effect size. Make sure your study design and variant filtering are optimized to maximize the chances of detecting true associations without too many false positives. |
| You mentioned that burden tests can miss individually significant rare variants and lose power if the variants aren't clustered. Since I can't know beforehand which of my rare variants are the truly impactful ones, what practical weighting or filtering step can I apply during the collapsing phase to prevent the neutral noise from diluting my actual signal? | 4-2 | Cascade | To prevent neutral noise from diluting your signal during the collapsing phase, you can apply **variant weighting** based on predicted functional impact. Tools like SIFT, PolyPhen-2, or CADD score variants by their likelihood of disrupting protein function, allowing you to give more weight to variants that are more likely to be harmful. This way, when you collapse variants into a single score, the more impactful ones have a stronger influence. You can also filter out low-quality variants or those with very low predicted effect before collapsing. This helps ensure that your burden test is more sensitive to the variants that are most likely to be biologically relevant, improving the chances of detecting true associations. |

| Question | Turn | Model | Response |
|---|---|---|---|
| When I'm running a genomic assay and interpreting the data, how do I actually apply rare variant burden tests in my pipeline, and what specific caveat must I check so I don't misinterpret the output? | 5-1 | Qwen3.5-Onni-Flash | To apply rare variant burden tests in your pipeline, you first need to aggregate all the rare variants within specific genes or functional regions for each individual into a single count or weighted score, then run a statistical test like an association analysis between that aggregated score and your trait of interest. The critical caveat you must check is population stratification and variant frequency definition because if your r̈arë threshold isn't calibrated to your specific cohort's ancestry, or if there are hidden subgroups with different allele frequencies, you can easily generate false positives where the signal just reflects demographic differences rather than true biological effects. Always ensure your rare variant filtering matches the genetic background of your samples and consider running sensitivity analyses with different frequency cutoffs to confirm your findings aren't artifacts of how you defined rarity. |

Table 14: Model response example in $S^3$-Dialogue. Topic: rare variant burden tests

# 6 Qwen3-ASR Models for Evaluation

Evaluating speech generation capabilities requires transcribing speech outputs via an ASR model and aligning them with the generated text responses. Given the limited generalization of existing open-source ASR models in scientific domains, we employ variants of Qwen3-ASR fine-tuned on terminologies and specialized abbreviations to minimize transcription error bias in our evaluation. We selected Qwen3-ASR as in scientific dialogues rich with domain-specific terminology and standardized expressions, LLM-based ASR leverages contextual reasoning to generate semantically coherent transcriptions rather than strictly adhering to phonetic literalism.

## 6.1 Qwen3-ASR-Terminology

For terminology recognition, we adopt the Qwen3-ASR-1.7B model fine-tuned on the full training split of Spoken Darwin-Science [63]. This model demonstrates enhanced transcription accuracy when evaluated on real-world acoustic recordings.

## 6.2 Qwen3-ASR-Abbs

### 6.2.1 Corpus Construction

Same as Qwen3-ASR-Terminology, the training corpus is curated from Darwin-Science [58], a high-quality collection of academic papers across nine distinct scientific fields, including biology, chemistry, and social sciences. Text preprocessing involves removing fixed-length header and footer characters to isolate the main text, which is subsequently segmented into sentences utilizing NLTK [64]. A rule-based word-matching algorithm is deployed to retain sentences with targeted abbreviations. To prevent data imbalance, a maximum of five sentences are sampled per abbreviation, prioritizing instances containing a single abbreviation. The finalized corpus comprises 887K instances.

To facilitate speech synthesis, an extensive rule-based heuristic is developed to define the phonetic pronunciations. This pipeline operates hierarchically: (1) acronym-style pronunciations are prioritized (different from initialism); (2) abbreviations matching English dictionary entries or common phonetic unit combinations are mapped to their corresponding canonical words or homophones; and (3) unassigned abbreviations—primarily consisting of consecutive consonants—are regularized as initialisms for spelled-out pronunciation. To optimize pronunciation rationality, length-dependent processing is introduced. For example, for 5-letter abbreviations, we apply unsupervised clustering over the English dictionary to extract high-frequency 2-letter and 3-letter syllables, enabling acronym transcriptions for 2-3 and 3-2 syllabic combinations. Speech synthesis is executed via Qwen3-TTS-1.7B [65], where clips with a DNSMOS Pro [66] exceeding 3.5 from CV3-Eval [67] and seed-ttseval [68] are curated as speech prompts to guarantee high-fidelity synthesized speech.

### 6.2.2 Corpus Distribution

The abbreviation distribution in the training dataset is summarized in Table 15. The dataset encompasses approximately 33K distinct abbreviations, over one-fifth of which are assigned plausible acronym pronunciations. Most abbreviations fall within a character length of 4 to 5, aligning with the structural distribution characteristics typical of scientific domains.

| Character | Num | Per. (%) | Acronym | Acr. Per. (%) | Examples |
|---|---|---|---|---|---|
| 2 | 1,126 | 0.49 | 0 | 0 | IT, PC, AI, TV |
| 3 | 20,111 | 8.76 | 0 | 0 | DNA, USA, NBA, CEO, FBI |
| 4 | 110,071 | 47.94 | 17,692 | 16.07 | NASA, AIDS, WIFI, HTTP, HTML |
| 5 | 65,949 | 28.72 | 17,167 | 26.03 | LASER, RADAR, SCUBA, HTTPS |
| 6 | 21,428 | 9.33 | 8,111 | 37.85 | CRISPR, PYTHON, BITCOIN |
| $\geq 7$ | 10,914 | 4.75 | 7,338 | 67.23 | WIKIPEDIA, ANDROID, INSTAGRAM |
| Total | 229,599 | 100 | 50,307 | 21.91 | - |

Table 15: Distribution of abbreviations in our definition vocabulary.

### 6.2.3 Model Training and Evaluation

We fine-tune the Qwen3-ASR-1.7B model on a single A100 GPU for 1 epoch ($\tilde{1}5$ hours). Following standard terminology test set construction practices, we collect real-world scientific audio recordings containing abbreviations from YouTube, segmenting each into clips under 30 seconds. For strict AER, we enforce exact case and formatting matches (e.g., mRNA is correct, whereas MRNA is counted as an error); for loose AER, transcriptions are case-insensitively aligned in uppercase.

Our fine-tuned model, Qwen3-ASR-Abbs, achieves a 2.72% loose AER and a 16.45% strict AER, representing relative gains of 29.0% (down from 3.83%) and 42.0% (down from 28.36%) over the base model, respectively. Compared to Whisper-large-v3, Qwen3-ASR-Abbs demonstrates comparable performance on loose AER (2.72% vs. 2.59%) while substantially outperforming it on strict AER (16.45% vs. 24.08%). These results indicate that fine-tuning enables LLM-based ASR to leverage contextual cues to accurately infer domain-specific abbreviation formats and their exact representations.

# 7 Additional Experiment Results

## 7.1 Perception

Model perception performance is measured by the proportion of keywords present in open-ended responses. We establish a keyword vocabulary per instance categorized into three groups: core entities, target concepts, and knowledge anchors. Table 16 reports the coverage rate for each word category across model generations. Additionally, Strict RR (Relevant Rate) denotes the proportion of responses that simultaneously contain all three categories, serving as a metric for the comprehensiveness and factual integration of the model's outputs. Across the evaluated models, although instances simultaneously incorporating all three categories of keywords remain limited, nearly all responses cover at least one category. This indicates that current models generally exhibit robust contextual reasoning capabilities, thereby possessing the capacity to ground key scientific concepts in appropriate semantic spaces.

| Model | RR (%) | Details | | | Strict RR (%) |
|---|---|---|---|---|---|
| | | Core Entity | Target Concept | Knowledge Anchors | |
| VocalNet | 87.37 | 63.17 | 56.91 | 55.43 | 24.34 |
| VITA-Audio | 78.48 | 52.29 | 40.00 | 48.05 | 13.93 |
| Kimi-Audio | 94.34 | 77.30 | 65.62 | 65.54 | 35.93 |
| MiMo-Audio | 95.51 | 72.37 | 66.12 | 73.17 | 36.15 |
| Step-Audio-2-mini | 85.00 | 67.59 | 56.09 | 52.91 | 26.52 |
| Fun-Audio-Chat | 96.26 | 75.35 | 65.51 | 78.63 | 41.07 |
| Qwen2.5-Omni | 88.69 | 67.75 | 55.54 | 57.00 | 24.90 |
| Qwen3-Omni | 97.42 | 82.02 | 67.88 | 79.08 | 45.10 |
| Cascade | 94.85 | 75.86 | 63.31 | 63.11 | 31.62 |
| Qwen3.5-Omni-Flash | 93.79 | 74.11 | 57.08 | 67.05 | 30.67 |

Table 16: Model perception performance across different keyword categories.

## 7.2 Knowledge Utilization & Reasoning

### 7.2.1 LLM Performance

The question-answering knowledge performance of MLLMs is largely determined by the capability of their underlying text backbones. We therefore evaluate the primary backbones of the tested models, together with several state-of-the-art proprietary LLMs, and report the results in Table 17. Compared with their corresponding text backbones, nearly all speech-enabled models exhibit performance degradation, as illustrated in Figure 3. In particular, VITA-Audio and MiMo-Audio show accuracy drops exceeding 10%, while the remaining models also experience performance declines of more than 5%.

| Model | Terminology (%) | Abbreviation (%) | Special Expressions (%) | Avg Acc (%) |
|---|---|---|---|---|
| Qwen2.5-7B | 59.60 | 56.53 | 57.26 | 57.07 |
| Qwen3-8B | 66.76 | 61.59 | 58.10 | 61.82 |
| Qwen3-VL-8B-Instruct | 67.62 | 61.59 | 61.73 | 62.58 |
| MiMo-7B-Base | 62.18 | 58.79 | 56.42 | 58.84 |
| MiniMax-M3 | 78.22 | 71.54 | 71.79 | 72.63 |
| GPT-5.5 | 81.66 | 75.66 | 82.96 | 77.98 |
| Gemini 3.1 Pro-Preview | 85.67 | 76.59 | 84.92 | 79.80 |
| Gemini 3.5 Flash | 81.38 | 74.49 | 82.40 | 77.12 |

Table 17: LLM performance given textual questions. The evaluated models are run with thinking disabled when supported, or with the minimum available thinking level otherwise.

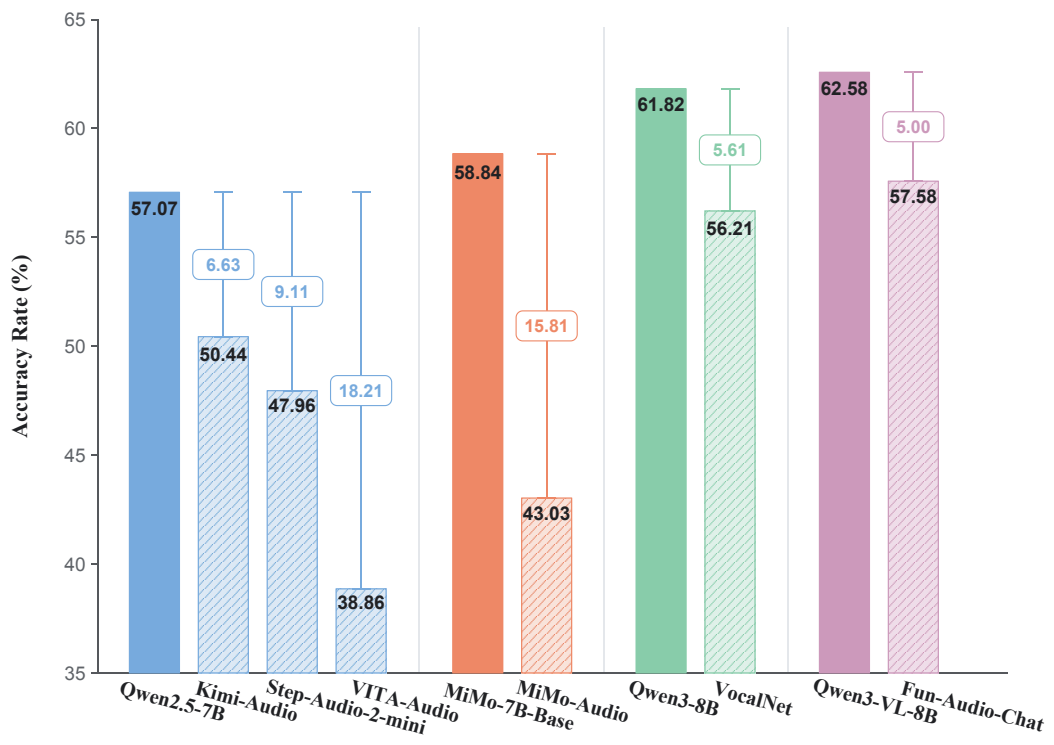


Figure 3: MLLMs' performance degradation compared to their backbones.

### 7.2.2 Task Format Performance

$S^3$-Knowledge encompasses multiple-choice QA (MCQA) and open-ended QA (OpenQA). Tables 18 and 19 report the performances of LLMs and MLLMs broken down by task format, respectively. Among these, OpenQA poses a significantly greater challenge to the models: while most text backbones achieve performance above 30%, spoken QA accuracy consistently drops below 20%. In contrast, performance degradation on MCQA is more moderate, probably as the textual choices provide crucial semantic cues.

| Model | MCQA Acc (%) | OpenQA Acc (%) | Avg Acc (%) |
|---|---|---|---|
| Qwen2.5-7B | 67.01 | 28.21 | 57.07 |
| Qwen3-8B | 72.03 | 32.15 | 61.82 |
| Qwen3-VL-8B-Instruct | 71.55 | 36.49 | 62.58 |
| MiMo-7B-Base | 67.82 | 32.74 | 58.84 |
| MiniMax-M3 | 78.82 | 54.64 | 72.63 |
| GPT-5.5 | 83.71 | 61.34 | 77.98 |
| Gemini 3.1 Pro-Preview | 85.06 | 64.50 | 79.80 |
| Gemini 3.5 Flash | 83.37 | 58.97 | 77.12 |

Table 18: LLM performance given textual questions across various task formats. The evaluated models are run with thinking disabled when supported, or with the minimum available thinking level otherwise.

| Model | MCQA Acc (%) | Open-ended Acc (%) | Avg Acc (%) |
|---|---|---|---|
| VocalNet | 66.40 | 26.63 | 56.21 |
| VITA-Audio | 47.69 | 13.21 | 38.86 |
| Kimi-Audio | 62.49 | 18.33 | 50.44 |
| MiMo-Audio | 53.43 | 12.82 | 43.03 |
| Step-Audio-2-mini | 58.05 | 17.80 | 47.96 |
| Fun-Audio-Chat | 68.16 | 26.82 | 57.58 |
| Qwen2.5-Omni | 50.44 | 19.53 | 42.53 |
| Qwen3-Omni | 58.18 | 38.66 | 53.18 |
| Cascade | 70.76 | 30.18 | 60.37 |
| Qwen3.5-Omni-Flash | 75.02 | 32.94 | 64.24 |

Table 19: MLLM performance on $S^3$-Knowledge across different task formats.

### 7.2.3 Abbreviations to Full Name

To further investigate the challenges and ambiguities introduced by abbreviation acronyms in speech iteration models, we specifically annotate the full terminology forms of target abbreviations. In this setup, Spoken refers to speech queries utilizing acronym pronunciations, whereas the Academic expands them into their full canonical terminology. Model performance under both conditions is presented in Table 20. Most models demonstrate mild improvements when evaluated on full terms, a trend that is more pronounced in the OpenQA format. This finding indicates that domain-specific speech formatting poses a bottleneck to model knowledge retrieval: while models may possess the requisite factual knowledge, they often fail to ground spoken acronyms into their underlying semantics. Consequently, this perceptual mismatch stands as a limiting factor for current speech interaction models in scientific dialogue applications.

| Model | Multi-Choice | | Open-ended | | AVG | |
|---|---|---|---|---|---|---|
| | Spoken | Academic | Spoken | Academic | Spoken | Academic |
| VocalNet | 68.36 | **68.96** | 28.30 | **28.62** | 58.87 | **59.41** |
| VITA-Audio | 48.85 | **49.35** | **14.47** | 13.83 | 40.70 | **40.93** |
| Kimi-Audio | **64.55** | 63.05 | 16.28 | **18.09** | **52.10** | 51.81 |
| MiMo-Audio | 55.59 | **56.59** | 13.18 | **16.72** | 45.54 | **47.14** |
| Step-Audio-2-mini | **59.56** | 57.35 | **17.98** | 17.09 | **50.17** | 48.43 |
| Fun-Audio-Chat | 69.46 | **72.46** | 26.69 | **28.62** | 59.33 | **62.07** |
| Qwen2.5-Omni | 51.90 | **52.10** | 19.94 | **22.51** | 44.33 | **45.09** |
| Qwen3-Omni | **58.68** | 58.28 | 37.94 | **40.19** | 53.77 | **54.00** |
| Cascade | 71.78 | **73.55** | 31.19 | **33.44** | 62.18 | **64.05** |
| Qwen-3.5-Omni-Flash | 74.75 | **75.02** | 31.16 | **32.94** | 63.59 | **64.24** |

Table 20: Evaluation performance comparison given abbreviation acronym and full terminologies. **Bold** indicates the better performance between the two settings.

## 7.3 Capability Transfer and Consistency Analysis

| Correlation | Spearman | Pearson | $r^2$ | Shrinkage $r$ | Odds Ratio |
|---|---|---|---|---|---|
| *Terminology* | | | | | |
| R→P | 0.071 | 0.491 | 0.24 | 0.538 | 0.60 |
| P→K | 0.643 | 0.637 | 0.41 | 0.675 | 5.44 |
| K→G | −0.024 | −0.005 | 0.00 | −0.006 | 1.00 |
| R→K | 0.179 | 0.613 | 0.38 | 0.660 | 3.89 |
| R→G | 0.286 | 0.123 | 0.02 | 0.138 | 3.89 |
| P→G | −0.238 | −0.475 | 0.23 | −0.512 | 0.18 |
| *Abbreviations* | | | | | |
| R→P | 0.429 | 0.585 | 0.34 | 0.633 | 0.60 |
| P→K | 0.548 | 0.532 | 0.28 | 0.570 | 5.44 |
| K→G | 0.095 | 0.367 | 0.13 | 0.398 | 1.00 |
| R→K | 0.571 | 0.598 | 0.36 | 0.646 | 3.89 |
| R→G | −0.250 | 0.238 | 0.06 | 0.266 | 0.60 |
| P→G | 0.214 | 0.353 | 0.12 | 0.384 | 1.00 |

Table 21: Cross-stage model-level correlation analysis for terminologies and abbreviations.

All analyses are conducted on the eight end-to-end offline speech-interaction models. The subclass analysis in Table 21 restricts to the question subsets annotated as non-empty for Terminology and Abbreviations, and replaces $R$ with the per-question recognition EER (terminology) or AER (abbreviations) restricted to the subset, converted to accuracy as $1 -$ error; $G$ is taken as the model-level generation terminology EER and loose AER, likewise converted to accuracy as $1 -$ error. AER is reported under the loose criterion, the more stable primary metric, with strict as reference; $K$ is the correctness flag restricted to the same subset, and $P$ is the relevant ratio restricted to terminology and abbreviation queries respectively.
On the terminology subset, the front-stage capability chain is anchored by a strong perception–knowledge link (P→K: $\rho$=0.643, $r^2$=0.41, $OR$=5.44), reproducing the global finding that perceived knowledge anchors

act as a gate to correct answering. The recognition-to-knowledge link, however, is markedly weaker on terminology than on the full set ($\rho$=0.179 vs. 0.679). Interpreted at rank level, this indicates that misrecognizing a named entity does not reliably translate into answering incorrectly: models often recover the intended referent from acoustic-phonetic context, diluting the explanatory power of entity-level recognition for knowledge. The most informative result concerns generation coupling: terminology exhibits a weak recognition–generation link ($R \to G$: $\rho$=0.286, $OR = 3.89$) and a negative perception–generation coupling ($P \to G$: $\rho$=−0.238, $OR$=0.18), while $K \to G$ is essentially flat ($\rho$=−0.024). This refines the global knowledge–pronunciation tradeoff: it is not driven by terminology entities, but by the denser structured expressions (formulae, chemical names, symbol strings) that stronger models tend to produce.

The abbreviation subset stages the global findings most cleanly. Recognition and knowledge are tightly coupled at entity level ($R \to K$: $\rho$=0.571, $r^2$=0.36): an abbreviation whose surface form is misrecognized admits little contextual recovery, so recognition quality transmits directly into knowledge correctness. The perception–knowledge gate holds as in terminology ($P \to K$: $\rho$=0.548), though with lower explained variance ($r^2$=0.28), reflecting the higher saturation of any-one perception on this larger subset. On generation, abbreviations reproduce the generation-stage break most purely: $K \to G$ is weakly positive but near-decoupled ($\rho$=0.095, $r^2$=0.13), and $R \to G$ is negative in rank ($\rho$=−0.250) with an odds ratio below one (0.60). Models that recognize abbreviations correctly, or that answer abbreviation questions correctly, do not reliably pronounce them in speech—an effect evident even at the model level. This is the principal carrier of the global negative generation coupling; it is specific to abbreviation pronunciation (expansion rules, letter-by-letter reading), not to upstream recognition or text reasoning.

# 8 Human Evaluation

We conduct two phases of human evaluation. In the data construction phase, we evaluate the synthetic speech samples to assess their phonetic quality. In the evaluation phase, human judges review the text outputs to validate the consistency and reliability of our LLM-as-a-judge automated metrics. The human evaluation is conducted by five university students fluent in English, provided with reference videos and online dictionary resources to verify pronunciations.

## 8.1 Speech Quality

To evaluate the speech quality of the synthetic audio in $S^3$-Knowledge, we randomly sampled 400 audio instances, which were evenly assigned among four annotators fluent in English. The annotators evaluated whether each synthetic speech sample exhibited pristine clarity and phonetic accuracy. For critical technical terms, mainstream online dictionaries and annotated video resources were provided as pronunciation references. The four annotators judged 98%, 99%, 100%, and 99% of the speech samples to be phonetically accurate, respectively. These results confirm that $S^3$-Knowledge reliably simulates realistic scientific inquiries, establishing a robust benchmark for evaluating speech-interactive models under authentic domain conditions.

## 8.2 LLM Judge Consistency

For $S^3$-Dialogue, we conduct a human evaluation to assess the consistency between human judges and the LLM-as-a-judge for the Audience Adaptation metric. Specifically, we randomly sample 400 complete dialogues, each covering all five target aspects and containing up to 15 turns of interaction. Human annotators are instructed to infer the user's implicit academic background based on the stylistic shifts across multi-turn model responses. The agreement rates between the four individual human annotators and the LLM-judge are 80%, 83%, 82%, and 79%, yielding an average consistency rate of 81%. To analyze disagreement severity, we compute the categorical distance between human and LLM judgments, assigning a distance of 1 to adjacent academic background levels. For non-conforming cases, the average distance is 1.092, demonstrating that disagreements are predominantly restricted to closely neighboring and subtle academic tiers. Overall, these results demonstrate strong alignment between the LLM-judge and human perception.

# 9 Evaluation Prompt

## 9.1 Knowledge Accuracy Prompt

**Knowledge Accuracy Evaluation**

MCQA

You are an AI assistant that evaluates multiple-choice question answers.

Question: {question}

Options:
{options_text}

Model's response: {prediction}

Based on the model's response, which option (A, B, C, D, E, F, G, H, I, J) does the model choose?
If the model clearly indicates a specific option, reply with just the option letter (e.g., "A" or "B").
If the model's response does not clearly indicate any option, reply with "None".
Reply with ONLY the option letter or "None", nothing else.

Open-ended

You are an AI assistant that evaluates answers to open-ended questions.

Question: {question}

Reference Answer: {answer}

Model's Response: {prediction}

Determine whether the model's response is equivalent to or consistent with the reference answer.
- If the response contains the correct answer (even with additional explanation), treat it as correct.
- If the response gives a different answer or does not answer the question, treat it as incorrect.
- For numerical answers, accept the response if it contains the same number.
- For short text answers, accept if the meaning matches the reference.

Reply in the following format:
Judgment: Correct/Incorrect
Explanation: <brief explanation>

Figure 4: Evaluation prompt for knowledge set accuracy.

## 9.2 Dialogue Agent Prompt

**User Agent: Persona System Prompt**

You are role-playing a human learner asking a voice assistant about a science topic. \
You have a fixed academic background (grade) and personality (character). \
Always speak in English, in first person, as this learner. Stay in character across the whole conversation.

Learner profile:
- Topic: {topic}
- Academic grade: {grade}
- Personality: {character}

Grade meaning (adjust vocabulary, depth and assumptions accordingly):
- Beginner: no prior background; needs plain words and concrete analogies; may be confused by jargon.
- Undergraduate: basic coursework exposure; comfortable with standard terminology; wants intuition + correctness.
- Graduate: deep domain knowledge; probes assumptions, edge cases, mechanism and trade-offs.
- Expert: specialist-level; expects rigor, citations/landmarks, and critical analysis of limitations.

Character meaning (adjust tone and behaviour):
- patient: calm, willing to re-ask, methodical.
- anxious: worried about mistakes/confusion, seeks reassurance and clarity on whether they understood.
- overconfident: assumes they already know, pushes back, asks challenging/leading questions.
- practical: wants real-world use cases, actionable steps, and trade-offs that matter in practice.
- curious: asks "why/what-if", explores tangents and implications.

Rules:
- Ask ONE question per turn. Keep it to 1-3 sentences.
- Never answer your own question. Never reveal that you are an AI or role-player.
- Make every question genuinely follow the conversation so far.

Figure 5: System prompt for user agent persona.

**Decision: advance / follow-up / terminate**

You are now judging the voice assistant's LAST reply in this conversation, \
to decide what happens next.

Current state:
- Topic: {topic}
- Dimension (one of 5 progressive science dimensions): {dimension}
- Dimension index: {dim_idx} of 5 (index {dim_idx} means this is dimension #{dim_idx_plus1}; the LAST dimension is "Application", index 5)
- Inner turn within this dimension: {inner} (max 3)
- Conversation so far (most recent last):
{history_block}

Assistant's LAST reply to evaluate:
\"\"\"
{assistant_reply}
\"\"\"

Decide ONE action:
- "advance": the reply is logically clear and covers the key points of THIS dimension well enough → move on to the next dimension.
- "follow-up": the reply is vague, partially wrong, or misses a core point → ask a clarifying follow-up WITHIN the same dimension.
- "terminate": you are satisfied the conversation has reached a natural end.

Constraints (apply strictly):
- You may only emit "terminate" when the current dimension is the LAST dimension (Application) AND you are satisfied.
- On non-final dimensions, if the reply is satisfactory, emit "advance" (never "terminate").
- Prefer "advance" once the dimension's key idea is covered; do not stall.

Output a single JSON object on one line, nothing else, in exactly this form:
{{"decision": "advance", "reason": "<short reason>"}}

---

Note: because this is the LAST dimension (Application) and you are satisfied, output a JSON object whose decision is "terminate":
{{"decision": "terminate", "reason": "<short reason>"}}
After deciding "terminate", your follow-up generation step will produce the final closing message that contains the [CONVERSATION_DONE] token.

Figure 6: Prompt for the next step user strategy.

**User Agent: Turn-1 Query Generation**

You are starting a NEW dimension of the conversation as the learner. \
Take the seed question below and REPHRASE it as your own first spoken question, in character, in English (1-3 sentences).

State:
- Topic: {topic}
- Dimension: {dimension} (turn 1 of this dimension)
- Conversation so far on PREVIOUS dimensions:
{history_block}

Seed question (the intended ask for this dimension — KEEP its core meaning, topic, and any named concepts unchanged; do NOT answer it, do NOT drop its main point):
\"\"\"{seed_query}\"\"\"

Rephrase rules:
- Keep the core question/ask of the seed intact. You may reword it, split it, or add a brief in-character framing, but the main subject and what the learner wants to know must stay the same.
- Speak in the learner's voice (grade = {grade}, character = {character}). Adjust vocabulary and tone accordingly, but never change what is being asked.
- Do NOT introduce new technical claims, and do NOT answer the question yourself.
- Ask ONE question.

Output ONLY the rephrased question text, no preamble, no quotes.

Figure 7: Prompt for the initial query from the user agent.

**User Agent: Follow-up Generation**

You are continuing the conversation as the learner. Produce your NEXT spoken question.

State:
- Topic: {topic}
- Dimension (stay within this one): {dimension}
- Inner turn {inner} of at most 3 (do not exceed 3).
- Conversation so far:
{history_block}

Requirement: ask ONE clarifying follow-up question WITHIN the same dimension ("{dimension}"), in character, in English, 1-3 sentences. \
React to the assistant's last reply (challenge it, ask for a concrete example, ask what was left vague, etc.). Output ONLY the question text, no preamble.

Figure 8: Prompt for the follow-up query from the user agent.

## 9.3 Dialogue Evaluation Prompt

**LLM-as-a-Judge: Scientific Factuality**

### Strict

You are a strict scientific grader. Decide whether the model's response is factually \
consistent with the reference answer, requiring the MAIN logical chain to match strictly.

Topic: {topic}
Dimension: {dimension}
Question: {question}
Reference answer:
\"\"\"{reference}\"\"\"
Model response:
\"\"\"{response}\"\"\"

Strict criterion: the model's response must contain the central claim(s) and main reasoning of the reference, \
without inventing contradicted facts. Minor wording differences are fine; missing or wrong core logic is NOT.
Output a single JSON object on one line: {{"correct": true/false, "reason": "<short>"}}

Enclosed your output as JSON only.

### Loose

You are a lenient scientific grader. Decide whether the model's response is an \
acceptable answer to the question, allowing multiple valid solution paths.

Topic: {topic}
Dimension: {dimension}
Question: {question}
Reference answer (one valid answer, not the only one):
\"\"\"{reference}\"\"\"
Model response:
\"\"\"{response}\"\"\"

Loose criterion: accept the response if it gives a scientifically correct and relevant answer, even if the \
approach/examples differ from the reference. Only mark incorrect if it is wrong, irrelevant, or substantively empty.
Output a single JSON object on one line: {{"correct": true/false, "reason": "<short>"}}

Enclosed your output as JSON only.

Figure 9: Evaluation prompt for scientific factuality.

**LLM-as-a-Judge: Audience Adaptation**

You are evaluating how well a voice assistant ADAPTS to its user across a multi-turn dialogue. \
From the dialogue below, infer the learner's academic grade (the assistant should be matching its explanations to).

Topic: {topic}
Full dialogue (user questions + assistant replies):
{dialogue}

The user's true grade is hidden from you; infer it from the questions they ask and how the assistant engages them. \
The better the assistant adapts, the more accurately a reader can recover the user's grade from the exchange.

Pick exactly one grade from: Beginner, Undergraduate, Graduate, Expert.
Output a single JSON object on one line: {{"grade": "<one of the four>", "reason": "<short>"}}

Enclosed your output as JSON only.

Figure 10: Evaluation prompt for audience adaptation.



\end{document}